\documentclass{article} 
\usepackage{iclr2027_conference,times}

\iclrpreprintcopy   

\usepackage{amsmath,amsfonts,bm}

\def\eqref#1{equation~\ref{#1}}

\def\1{\bm{1}}

\DeclareMathAlphabet{\mathsfit}{\encodingdefault}{\sfdefault}{m}{sl}
\SetMathAlphabet{\mathsfit}{bold}{\encodingdefault}{\sfdefault}{bx}{n}

\usepackage{hyperref}
\usepackage{url}

\usepackage{multirow}
\usepackage{titlesec}
\usepackage{titletoc}
\usepackage{wrapfig}
\usepackage{tcolorbox}
\usepackage{enumitem}

\usepackage{algorithm}
\usepackage{algpseudocode}

\usepackage{placeins}
\usepackage{booktabs}

\usepackage{multirow}
\usepackage{titlesec}
\usepackage{titletoc}
\usepackage{wrapfig}

\graphicspath{{figures/}}

\definecolor{copper}{rgb}{0.82, 0.38, 0.10}
\definecolor{softrose}{rgb}{0.90, 0.40, 0.55}
\hypersetup{
    colorlinks,
    linkcolor={blue},
    citecolor=copper,
    urlcolor={blue}
}

\title{Temporal Recurrence Favors Fewer Layers}

\author{%
Ivan Anokhin\textsuperscript{1,2,3}\quad
Johan Obando-Ceron\textsuperscript{1,2}\quad
Irina Rish\textsuperscript{1,2}\quad
Sebastian Risi\textsuperscript{3}\\[1em]
{\normalfont\small
\textsuperscript{1}Mila -- Qu\'ebec AI Institute\quad
\textsuperscript{2}Universit\'e de Montr\'eal\quad 
\textsuperscript{3}Sakana AI 
}\\
\texttt{\{ivan.anokhin,johan.ceron,irina.rish\}@mila.quebec}\\
\texttt{sebastianrisi@sakana.ai}
}

\begin{document}

\maketitle

\begin{abstract}
In streaming tasks, recurrent models can carry latent computation across time, allowing each update to build on representations produced earlier. 
This raises a basic question: once temporal recurrence provides sequential computation across steps, how much depth is still needed within each step? 
Prior work has shown that recurrence can make shallow models competitive.
We instead study this question as a compute-allocation problem, varying within-step depth, expert width, and the number of parallel experts per layer across several compute budgets. 
For each budget, we compare the best observed recurrent and non-recurrent allocations and the performance they achieve under approximately matched per-step computation. 
Across Sokoban and autoregressive FineWeb language modeling, we find that temporal recurrence shifts the best observed compute allocation toward substantially fewer layers, with comparable or better performance.
\end{abstract}

\begingroup
\renewcommand{\thefootnote}{}
\footnotetext{Our code is available at
\href{https://github.com/avecplezir/fewer-layers-with-rec.git}{fewer-layers-with-rec}.}
\endgroup

\section{Introduction}

Across many deep learning tasks, models benefit from progressively refining their internal representations through a sequence of latent transformations~\citep{he2016deep, brown2020language,wang20261000,dehghani2018universal}. This refinement is commonly implemented by stacking layers, allowing later layers to build on intermediate states produced by earlier ones. 
Depth therefore controls one important source of sequential computation available to a model.

In natural \textit{streaming} tasks, sequential computation can also accumulate across time. The model is repeatedly applied as the task unfolds, for example, across tokens in language modeling, frames in video, or observation steps in reinforcement learning. A recurrent model can carry a latent state across these steps, allowing each update to build on computations performed at earlier steps. Recurrence can therefore do more than preserve information from the past: it can also continue refining representations or plans for future predictions and actions. 
This temporal recurrence is different from depth recurrence, as used in looped Transformers, where shared blocks are repeatedly applied  on the same input before producing an output~\citep{dehghani2018universal,darlow2025continuous,jolicoeurmartineau2025trm,geiping2026scaling}. Here, computation continues across successive token or environment steps as new inputs arrive. 

Whether sequential computation can accumulate across external steps depends on the structure of the task. It can do so when several consecutive steps contribute to the same broader problem, allowing computation performed at one step to remain useful later. 
One favorable setting is when relatively easy predictions or actions come before a harder one: an agent may refine a plan while walking through a corridor, and a language model may continue refining its representation of a passage while generating locally predictable tokens. In contrast, if each step presents a new independent problem that must be solved immediately, earlier computation has no value for later steps.

Prior work demonstrates that models can learn useful computation across external steps. State-tracking studies train models to predict the current state after each input operation~\citep{merrill2024illusion}. 
On difficult permutation-composition tasks, single-layer recurrent models succeed across the tested sequence lengths, whereas Transformers require increasing depth to maintain accuracy. 
Evidence also comes from reinforcement learning, where recurrent agents learn planning-like behavior in fully observed Sokoban environments \citep{guez2019investigation}. Subsequent analysis shows that their internal plans develop across environment steps and causally influence their actions \citep{bush2025interpreting}. 

This raises the central question of the paper: \textit{How much within-step depth is still needed once temporal recurrence is available?} 
Prior work has shown that recurrence can make shallow models competitive~\citep{merrill2024illusion,guez2019investigation,fan2020addressing,oncescu2026recurrent}.
We instead formulate this question as a compute-allocation problem, comparing similar recurrent and non-recurrent models. 
Let \(S\) denote the number of sequential stages within that step, and let \(P\) denote the parallel work performed at each stage. The total work per step is then approximately $W \approx S \cdot P.$ At fixed $W$, allocations with fewer sequential stages spend more work on parallel computation at each stage. We ask whether temporal recurrence changes the best allocation: do recurrent models require a smaller \(S\) than non-recurrent models to achieve comparable performance? 

\begin{figure}[h]
    \centering
    \includegraphics[width=0.98\linewidth]{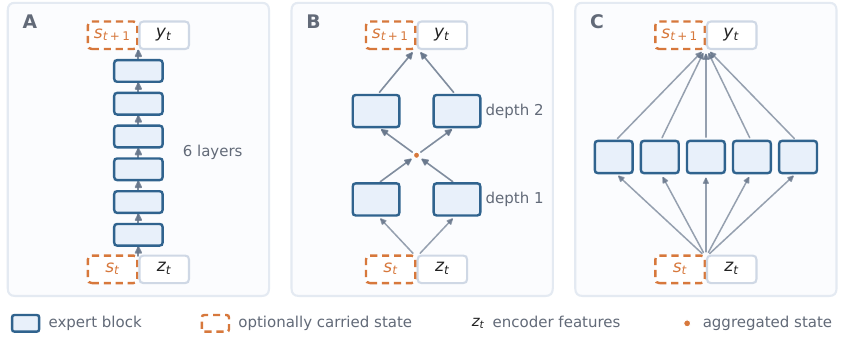}
    \vspace{-0.1cm}
\caption{ \textbf{Three example allocations under the same compute budget.}
The three example architectures schematically use the same approximate work budget but allocate it differently. \textbf{(A)} applies six expert blocks sequentially. \textbf{(B)} uses two depth levels, with two wider experts evaluated in parallel at each level. \textbf{(C)} evaluates five experts in parallel within a single depth level. Dashed boxes labeled \(s_t\) and \(s_{t+1}\) denote latent state that may be carried between external steps in the recurrent variant. Expert-block area represents approximate computational cost, and the total expert-block area is the same across three examples.}
    \label{fig:serial-to-parallel-experts}
\vspace{-0.5cm}
\end{figure}

To realize this comparison, we use recurrent and non-recurrent models with the same three allocation axes, illustrated in Fig.~\ref{fig:serial-to-parallel-experts}. 
The within-step depth $L$ controls the number of successive stages (layers) $S$. Within each stage, multiple experts are evaluated in parallel and their outputs are combined, allowing \(P\) to be increased in two ways: by adding more experts \(E\), or by increasing the hidden dimension \(d\) of each expert. The recurrent variant carries a latent state across external steps, while the non-recurrent variant does not. Varying \(L\), \(E\), and \(d\) under matched work lets us compare how much within-step depth remains useful with and without temporal recurrence. 

Our experiments support two main conclusions:

\begin{itemize}[leftmargin=*,nosep]
    \item \textbf{Temporal recurrence shifts the best compute allocation toward fewer layers.} Across Sokoban and streamed FineWeb, the best observed recurrent allocations use substantially fewer layers per step than the best non-recurrent allocations, while achieving comparable or better performance under approximately matched per-step work.

    \item \textbf{Recurrence and parallel capacity play different roles.} Increasing per-layer expert count benefits both recurrent and non-recurrent models, while gains from additional layers saturate earlier with recurrence. Recurrence therefore reduces the performance cost of
    using fewer layers, while expert count and width provide useful ways to spend the reallocated compute.
\end{itemize}

More broadly, we hypothesize that temporal recurrence will shift the best compute allocation toward shallower within-step updates in other streaming settings where successive steps contribute to a shared problem and carried state can support continued refinement.

\section{Why Fewer Layers Matter}

The compute-allocation tradeoff between sequential stages $S$ and parallel work $P$ matters also for latency. Under a fixed work budget, shifting computation from $S$ into $P$ reduces the serial part of the update, potentially decreasing latency if enough hardware parallelism is available. The actual speedups also depend on implementation details, memory bandwidth, and communication overhead. 

This potential reduction in per-update latency is particularly important for general nonlinear RNNs, whose state updates must generally be unrolled sequentially across tokens during language-model pretraining. Because the latency of each update accumulates at every token in the sequence, deeper updates lengthen both the forward unroll and the backward pass through the sequence. If comparable performance can be preserved while reallocating work from sequential stages \(S\) to parallel work \(P\), reducing the latency of each update could improve training throughput and make large-scale pretraining of general nonlinear RNNs more practical.

Other sequence architectures address this scalability problem by changing how computation is organized across tokens. Transformers process sequence positions in parallel during training, while structured recurrent models, such as Mamba, constrain their state transitions so that they can be evaluated with parallel scans across the sequence \citep{vaswani2017attention,gu2021efficiently,gu2023mamba}. These gains in temporal parallelism come with an expressivity trade-off. Under standard precision assumptions and complexity-theoretic conjectures, theory places fixed-depth Transformers and parallelizable linear recurrent models in computational classes believed to be weaker than attainable by general nonlinear RNNs \citep{merrill2024illusion,merrill2026linear}. 
This trade-off motivates preserving general nonlinear recurrence and studying whether work can be reallocated from \(S\) to \(P\) without sacrificing performance, as a potential route to improving training throughput.

There are also pieces of evidence supporting the possibility of parallelization in general. 
Recent transformer studies suggest that some computations normally implemented as strict depth may be more exchangeable than the architecture implies. \citet{sun2025transformer} find that middle layers of pretrained transformers share substantial representational structure, and that some tasks tolerate skipping, reordering, or parallelizing layers with limited degradation. 
\citet{freiberger2024layershuffle} train vision transformers to tolerate randomized attention-layer execution order, showing that robustness to arbitrary layer order can be learned, though with an accuracy cost.

\section{Preliminaries}
\label{sec:preliminaries}

We conduct our study in two settings that support updating a latent state over long external sequences: deep reinforcement learning and autoregressive language modeling. In both cases, the model repeatedly receives new input, updates an internal state when recurrence is used, and produces an action or prediction. 


\paragraph{Deep Reinforcement Learning.} In reinforcement learning, an agent interacts with an environment over timesteps $t=1,2,\ldots$ \citep{sutton1998reinforcement,mnih2016asynchronous,schwarzer23a}. At each step it receives an observation $o_t$, updates an internal latent state $s_t$, selects an action $a_t \sim \pi(a_t \mid s_t)$, and receives a reward $r_t$. The goal is to maximize the expected discounted return,
$J(\pi) = \mathbb{E}_{\pi}\left[\sum_{t=1}^{\infty} \gamma^{t-1} r_t \right],$
where $\gamma \in [0,1)$ is a discount factor that determines the relative importance of immediate and future rewards. The policy $\pi(a_t \mid s_t)$ maps the current latent state to a distribution over actions, and the expectation is taken over trajectories induced by the policy and environment dynamics. Our experiments use actor--critic agents \citep{konda1999actor,mnih2016asynchronous}.

Recurrent architectures are commonly used to maintain belief-like latent states that summarize trajectory history in partially observable environments and support long-horizon decision making \citep{hausknecht2015deep,heess2015memory,wang2016learning,hefny2018recurrent}. In this work, we emphasize a complementary use of recurrent computation in reinforcement learning: recurrence can support reasoning-like latent refinement, not merely memory-like tracking. 

\paragraph{Autoregressive Sequence Modeling.} Autoregressive sequence modeling aims to predict each token from its preceding context. Given a token sequence $x_{1:T}$, the objective is to model $p(x_{1:T}) = \prod_{t=1}^{T} p(x_t \mid x_{<t})$, or equivalently to predict the next-token distribution at each position \citep{bengio2003neural,vaswani2017attention}. This requires representations that accumulate and refine contextual information over long token streams. 

Classical recurrent sequence models maintain a latent state across token updates \citep{hochreiter1997long} and update that state sequentially during training. Modern long-sequence architectures often designed to improve training efficiency by enabling parallel scanning computation over sequence positions \citep{vaswani2017attention,dai2019transformer,gu2021efficiently,gu2023mamba}. Our version of the recurrent Transformer is not designed to support parallel scanning; instead, it emphasizes expressive latent-state updates with feedback connections through the carried state. This is intended to allow computation to compose across successive token steps rather than being concentrated within a single deep feed-forward pass.

\section{Experimental Framework}
\label{sec:parallel-recurrent-experts}

We introduce the Parallel Experts framework for studying how compute should be allocated between sequential and parallel computation. 
It is designed to expose a clean tradeoff between within-step depth and parallel capacity while allowing latent state to be carried across external steps.
At each external step, several expert modules read the current input
and a shared latent state, compute candidate updates in parallel,
and merge their outputs into the shared state. In recurrent models,
the final state is carried to the next external step. This construction exposes three allocation axes: within-step depth \(L\), expert count \(E\) per depth level, and expert width \(d\).

The depth $L$ counts the number of sequential depth level within one step. The expert count $E$ counts experts evaluated in parallel at each depth level, and $d$ controls per-expert capacity by adjusting expert's hidden dimension (width). If $C(d)$ is the cost of one expert transformation, the approximated work is
\begin{equation}
 W \approx L \cdot E \cdot C(d).   
\end{equation}
We use this proxy to construct matched-budget allocation sweeps that are intended to compare where computation should be placed to maximize the performance for recurrent and non-recurrent architectures.

At each environment or sequence step, the model receives an external input \(x_t\). Let \(z_t=e(x_t)\) denote the encoded representation of this external input. At step \(t\), the model also receives an incoming shared latent state \(s_t^0\), expert-local activations \(h_t^{0,i}\), and architecture-specific memories \(m_{t-1}^{\ell,i}\), where \(\ell\in\{1,\ldots,L\}\) indexes depth levels and \(i\in\{1,\ldots,E\}\) indexes per depth level parallel experts.
Each depth level applies all experts in parallel:
\begin{equation}
\label{eq:update}
\begin{gathered}
    \begin{aligned}
        h_t^{\ell,i},\,m_t^{\ell,i}
        &= f_{\theta_{\ell,i}}\!\left(z_t, s_t^{\ell-1}, h_t^{\ell-1,i}, m_{t-1}^{\ell,i}\right),
        \qquad i=1,\ldots,E, \\
        s_t^\ell
        &= \frac{1}{\sqrt{E}}\sum_{i=1}^{E} h_t^{\ell,i},
        \qquad \ell=1,\ldots,L.
    \end{aligned} \\
\end{gathered}
\end{equation}
This normalized sum aggregates expert proposals while keeping the variance approximately the same as the number of contributing experts changes. 
To maximize the sequential computation available on the carried state,
the recurrent variant uses top-down feedback, feeding the final state
into the first depth level of the next step \citep{zilly2017recurrent,guez2019investigation,fan2020addressing}:
\begin{equation}
\begin{aligned}
    y_t &= g(s_t^L),
    \qquad 
    h_{t+1}^{0,i}=h_{t}^{L,i},
    \qquad
    s_{t+1}^0=s_t^L.
\end{aligned}
\end{equation}
Here \(y_t\) denotes action logits and a value estimate in reinforcement learning, or next-token logits in language modeling.

More specifically, in the LSTM instantiation \(m_t^{\ell,i}\) corresponds to the cell state \(c_t^{\ell,i}\), and in the transformer --- the key--value (KV) cache.
Full pseudocode and architecture-specific instantiations are given in Appendix~\ref{sec:appendix-architectures} and alternative aggregation rules are studied in Appendix~\ref{sec:appendix-aggregation-ablation}. 

The Parallel Experts framework is intended as a probe framework rather than a claim of architectural optimality. We use parallel experts because they expose a clean tradeoff between within-step depth and parallel capacity while optionally allowing a carried latent state. Other MoE mechanisms, including routed MLP experts~\citep{lepikhin2020gshard}, Soft MoE variants~\citep{puigcerver2023sparse}, or alternative recurrent communication schemes, may shift the quantitative frontier, but we expect the qualitative conclusion to stay the same. 

\section{Experiments}

Our experiments test different compute allocations. The Parallel Experts framework exposes three ways to spend compute: within-step depth $L$, per-layer expert count $E$, or width $d$ of each expert. 
The central question is whether, once recurrent state is carried across the external steps, the best allocations are still require deep sequential computation within each step.

For our main experiments we evaluate different allocations in two settings: Sokoban, using the \texttt{Sokoban-v0} environment from Jumanji~\citep{bonnet2024jumanji}, and streamed FineWeb language modeling~\citep{penedo2024fineweb}.
We report episodic return for reinforcement learning and validation
cross-entropy loss for language modeling. Unless stated otherwise, in RL each configuration is run with at least three random seeds.  
Additional implementation details are reported in Appendix~\ref{sec:appendix-hyperparameters}.

\subsection{Sokoban}
\label{sec:sokoban}

Sokoban is a fully observable puzzle environment in which an agent must push four boxes onto four target locations on a $10\times10$ grid.
Levels are procedurally generated, and mistakes can be irreversible: for example, pushing a box into a corner without a target makes the level unsolvable.
Planning is therefore important even though the agent observes the entire board at every step.
Prior work showed that DRC agents learn planning-like behavior in Sokoban \citep{guez2019investigation}, and later analysis found that their internal plans develop progressively through recurrence and causally affect their actions \citep{bush2025interpreting}.
These properties make Sokoban a natural setting for studying whether temporal recurrence reduces the within-step depth needed for planning.

\begin{figure*}[h]
    \vspace{-0.3cm}
    \centering
    \includegraphics[width=0.98\textwidth]{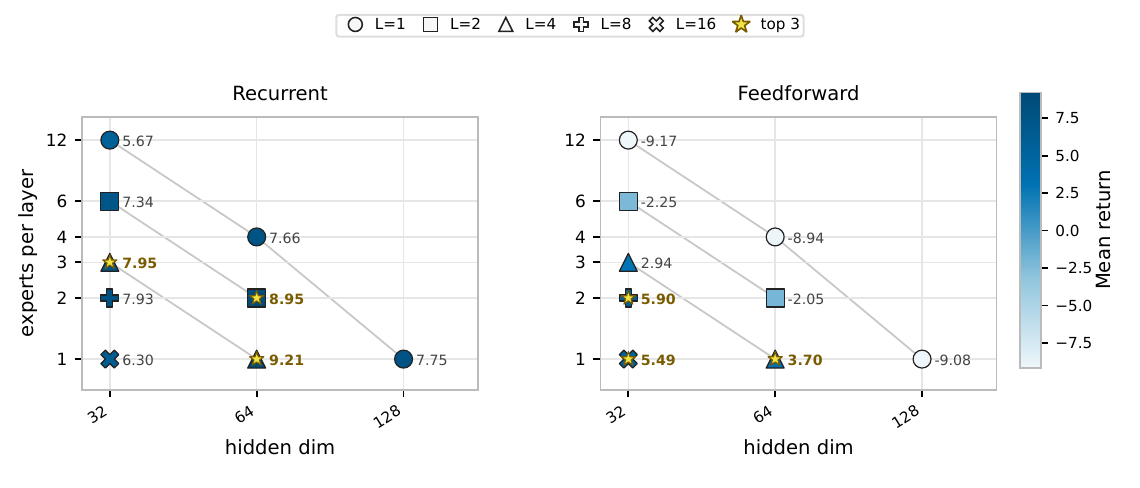}
    \vspace{-0.5cm}
    \caption{\textbf{Compute allocations in Sokoban under the small per-step budget.} 
    We compare recurrent and non-recurrent ConvLSTM agents across different compute allocations, varying within-step depth \(L\), per-layer expert count \(E\), and expert width \(d\) under approximately matched compute budget.
    Higher mean return is better. The best observed recurrent allocation uses four layers and achieves a higher return than the best non-recurrent allocation, which uses eight layers.}
    \label{fig:appendix-sokoban-small-lstm-recurrent-feedforward}
\end{figure*}

We use a DRC-inspired architecture with ConvLSTM experts \citep{guez2019investigation}, organized according to our Parallel Experts framework.
The original DRC setup combines temporal recurrence with depth-recurrence, repeated passes through the recurrent stack on the same observation.
To focus on temporal recurrence, we instead make a single pass through the $L$ depth levels per environment step, without additional internal looping.
We compare models trained with state carried across environment steps against models trained without carried state, keeping the remaining architecture and training setup the same.
For both variants, we vary within-step depth $L$, expert count $E$, and expert width $d$ under approximately matched per-step compute budgets.
This comparison tests whether temporal recurrence changes how compute is best allocated between within-step depth and parallel capacity.

Fig.~\ref{fig:appendix-sokoban-small-lstm-recurrent-feedforward} shows the allocation sweeps for both variants under the small compute budget.
With recurrence, the performance saturates at $L=4$, while the non-recurrent version saturates at depth 8 with much worse performance. 
Temporal recurrence changes not only the achieved performance but also the preferred compute allocation, favoring fewer sequential stages and more parallel capacity.

\begin{figure*}[t]
    \centering
    \includegraphics[width=0.98\textwidth]{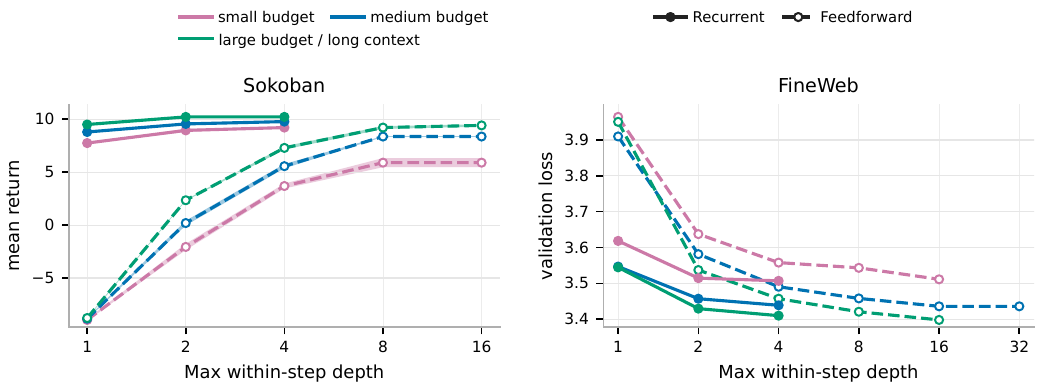}
    \vspace{-0.3cm}
    \caption{\textbf{Within-step depth saturation for recurrent and non-recurrent models.}
    We report the best result among tested allocations with $L$ or fewer within-step layers, within each compute budget.
    \textbf{Left}: Sokoban mean episodic return with one standard error across three seeds.
    \textbf{Right}: FineWeb validation loss, lower is better. Recurrent models reach their best observed performance with $2$--$4$ layers, while non-recurrent models benefit from greater depth.}
    \label{fig:fixed-compute-retained-improvement}
     \vspace{-0.3cm}
\end{figure*}

We repeat the allocation sweeps at medium and large compute budgets.
Fig.~\ref{fig:fixed-compute-retained-improvement} summarizes all three budgets, reporting the best observed
mean return over expert counts, widths, and depths up to $L$.
Across the tested budgets, recurrent performance saturates by depth 2-4, whereas non-recurrent models benefit from greater within-step depth, for example, still improving at a within-step depth of 16 under the large budget. 
At the same time, recurrent model performance is better across all settings, and larger budgets continue to improve both architectures at every tested within-step depth, showing that additional parallel compute remains a useful scaling axis across different configurations.
The full medium- and large-budget sweeps are reported in Appendix~\ref{sec:appendix-convlstm-fixed-compute-sweeps}.

\begin{wrapfigure}{r}{0.35924369747 \textwidth}
\vspace{-0.6cm}
\includegraphics[width=\linewidth]{
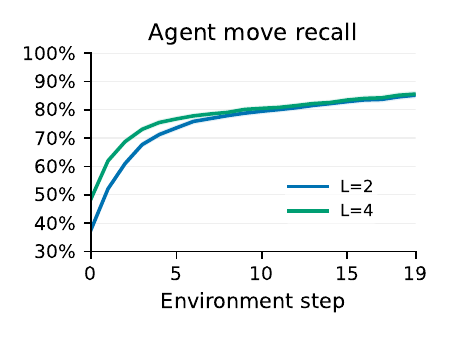}
\vspace{-0.8cm}
\caption{\textbf{Agent-move recall.} The four layer model initially has more complete plans, but the two layer
model catches up over the next few environment steps. Both models achieve almost identical performance.}
\vspace{-0.4cm}
\label{fig:sokoban-agent-move-recall}
\end{wrapfigure}

For a more direct architectural comparison, we also test ConvGLU models, whose experts apply a $3 \times 3$ convolution followed by GLU activation. These models omit the LSTM cell-state machinery retained in the non-recurrent ConvLSTM baseline.
Here, the only architectural difference is one extra projection in the recurrent variant, feeding the previous step's final residual stream into the first depth level.
We observe the same earlier depth saturation with recurrence under this single architectural intervention (see Appendix~\ref{sec:appendix-convglu-sokoban}).

To examine how planning develops at different within-step depths, we apply the spatial probes of \citet{bush2025interpreting} to the final residual representations of two large budget recurrent models with $L=2$ and $L=4$.
The probes predict the direction of the agent's movements in each spatial location. Fig.~\ref{fig:sokoban-agent-move-recall} compares recall across environment steps, with thresholds calibrated to approximately $93\%$ precision. The $L=4$ model initially has higher
recall, but the $L=2$ model catches up over the next few environment steps, after which both maintain similar recall. Both models also have almost identical performance. This pattern is consistent with the shallower model compensating for less planning within each update by continuing the computation across environment steps, catching up early enough to preserve task performance. Full probing details and additional depth comparisons are provided in
Appendix~\ref{sec:appendix-sokoban-agent-move-decoding}.

\subsection{Language Modeling}
\label{sec:language-modeling}

We next test whether the compute-allocation trends observed in Sokoban extend to language modeling. We train recurrent and non-recurrent models for next-token prediction on approximately one billion tokens from FineWeb and compare validation loss under approximately matched per-token compute budgets. We use Transformer blocks within our Parallel Experts framework, with the non-recurrent model reducing to a standard Transformer when $E=1$. The recurrent variant feeds the final residual stream into the first layer at the next token step, with additional adjustments for stability and memory efficiency described in Appendix~\ref{sec:appendix-architectures}. 


During training, recurrent models process continuous token streams, carrying their state and sliding window KV caches across training sequences and resetting them only at shard boundaries. Non-recurrent models are trained on independent training sequences, allowing parallel computation across token positions. We use fixed KV cache lengths of 128 and 1024 for recurrent models against non-recurrent training sequence lengths of $256$ and $2048$ tokens, respectively, approximately matching the average attention-context length during training. Both variants use the same total training-token budget and number of tokens per optimizer update. At evaluation, both variants process the validation stream continuously using sliding-window KV caches of the same length ($128$ or $1024$). Further streaming implementation details are provided in Appendix~\ref{sec:appendix-nanogpt-streaming}.

\begin{figure}[h]
    \vspace{-0.4cm}
    \centering
    \includegraphics[width=0.92\linewidth]{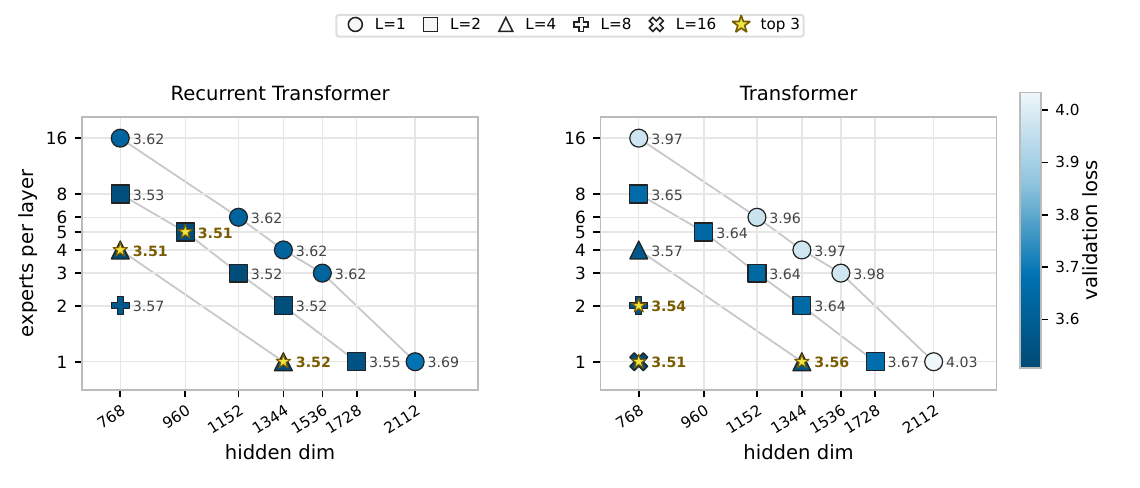}
    \vspace{-0.4cm}
    \caption{\textbf{Compute allocations in FineWeb under the small per-token budget.} We compare recurrent and non-recurrent Transformer allocations trained for next-token prediction on approximately one billion FineWeb tokens.
    Validation cross-entropy is shown; lower is better. Under approximately matched per-token work, the recurrent model reaches a validation loss of approximately \(3.51\) with two layers, comparable to the best observed non-recurrent allocation, which uses sixteen layers.}
    \label{fig:fineweb-small-budget}
\end{figure}

\paragraph{Results.} Fig.~\ref{fig:fineweb-small-budget} shows the allocation sweeps under the small compute budget with a $128$-token attention window. The recurrent model reaches a validation loss of approximately $3.51$ with only two depth levels ($L=2$), comparable to the best non-recurrent allocation at $L=16$. As in Sokoban, the preferred allocation shifts toward fewer sequential stages and greater parallel capacity when recurrence is available.

We extend the comparison to a medium compute budget and to the models trained and evaluated with an average $1024$-token attention window. Fig.~\ref{fig:fixed-compute-retained-improvement} summarizes these settings, reporting the lowest validation loss among allocations with depth at most $L$. Across the tested settings, recurrent models reach their best observed loss at $L=2$ or $4$, whereas non-recurrent models continue to improve up to $L=16$ or $32$. Although the preferred depth depends on the setting, recurrent models consistently achieve comparable validation loss with substantially less within-step depth.

\subsection{What Makes the Earlier Saturation Work?}

\begin{figure*}[h]
    \centering
    \includegraphics[width=0.98\linewidth]{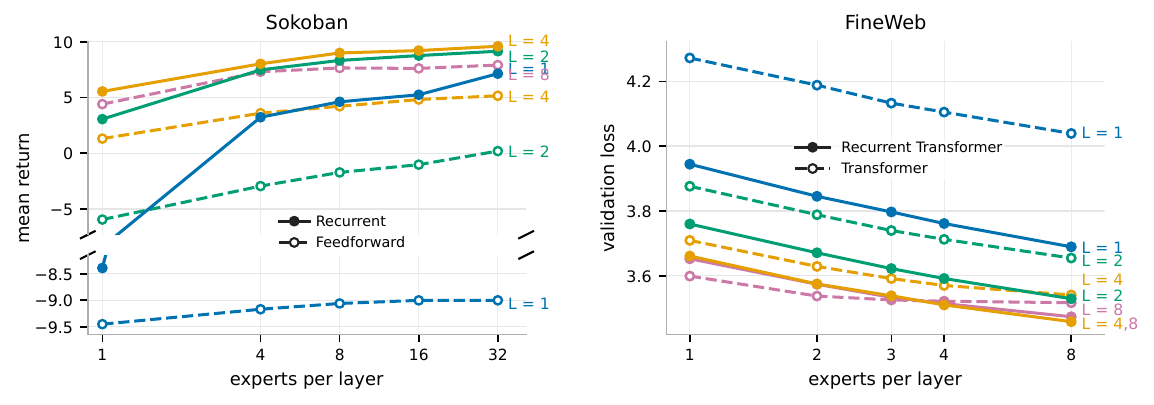}
    \vspace{-0.3cm}
\caption{\textbf{Scaling within-step depth and experts per layer.} Color denotes within-step depth $L$. Increasing experts per layer benefits both recurrent and non-recurrent models,
while gains from additional within-step depth saturate earlier with recurrence.
}
\label{fig:scale-feedforward-vs-recurrent}
\end{figure*}

The early saturation appears to reflect two approximately separable effects. In Fig.~\ref{fig:scale-feedforward-vs-recurrent}, increasing per-layer expert count improves both recurrent and non-recurrent models, suggesting that parallel capacity is broadly useful regardless of recurrence. In contrast, the depth-response curve depends strongly on recurrence: non-recurrent performance performance is much more sensitive to reducing $L$, whereas recurrent models saturate at substantially shallower within-step depth. Thus, recurrence primarily lowers the cost of reallocating compute away from depth, while experts/width provide a useful parallel place for that compute. Finally, parallel experts must communicate through shared state. In the communication-radius ablation (Appendix~\ref{sec:appendix-communication-radius}), Full shared-state access performs best while the narrower communication radii lower performance monotonically. Interestingly, Sokoban performance continues to improve beyond the eight-expert regime explored in prior deep RL Soft MoE work \citep{obando-ceron24b,sokar2025don}.

\section{Related Work}
\label{sec:related_work}

\paragraph{Temporal Recurrence and Model Depth.}
Several prior works provide evidence that temporal recurrence across external steps, such as token steps in streamed language modeling or environment steps in reinforcement learning, can make shallower recurrent neural networks more competitive with standard feed-forward architectures without temporal recurrence. Feedback Transformer~\citep{fan2020addressing} expose top-level representations from previous tokens to the current token's computation and show that such feedback makes shallow models with fewer parameters more competitive with standard Transformers that have more layers.

In reinforcement learning, DRC~\citep{guez2019investigation} carries recurrent state across steps of the Sokoban environment and outperforms deep feed-forward baselines. A closely related classical work, RHN~\citep{zilly2017recurrent}, studies how to scale within-step recurrent depth under a fixed parameter budget by trading off depth against width, with depth saturating at approximately 8--9 levels. Importantly, however, compute still grows substantially with depth in the RHN setup, even though the parameter count is held fixed. Similarly, recent recurrent Transformer work~\citep{oncescu2026recurrent} finds favorable depth--width tradeoffs under a fixed parameter budget. In contrast, our work keeps \emph{compute} approximately \emph{fixed}, rather than parameter count, and studies the tradeoff between within-step depth, expert width, and per-layer \emph{expert count} for recurrent and non-recurrent models.
In this sense, our work complements and extends prior studies.

\paragraph{Depth Recurrence, Mixture-of-Experts and Compute Allocation.}

A long line of work studies neural computation as an iterative refinement process in which representations are progressively updated through recurrent/looping transformations on the same input, rather than being computed in a single feed-forward pass. Universal Transformers~\citep{dehghani2018universal} reformulated Transformer layers as recurrent updates applied across depth, while Adaptive Computation Time~\citep{graves2016adaptive} and PonderNet~\citep{banino2021pondernet} explored dynamically allocating additional sequential computation to difficult inputs.

A few recent works have also combined depth recurrence with Mixture-of-Experts approaches, motivated by the lack of parameter capacity in these looping architectures; examples include recurrent modular architectures such as Recurrent Independent Mechanisms (RIMs)~\citep{goyal2019recurrent} and recurrent MoE systems such as MoEUT~\citep{321387ba}. Mixture-of-experts (MoE) architectures scale model capacity through conditional or parallel computation~\citep{jacobs1991adaptive,jordan1994hierarchical}. Modern sparse MoE systems demonstrate that parameter count can scale substantially while keeping per-token computation nearly constant~\citep{shazeer2017outrageously,lepikhin2020gshard,fedus2022switch,du2022glam}. Most prior work integrates experts inside fixed-depth feed-forward architectures, where the dominant scaling axis remains serial layer composition. Other works study improved routing and expert utilization, including differentiable aggregation mechanisms such as Soft MoE~\citep{puigcerver2023sparse,obando-ceron24b,willi2024mixture}. In our work, we focus mainly on temporal recurrence rather than depth recurrence, and MoE appears in our framework as a way to study a \emph{sequential-to-parallel compute allocation problem}. 
This perspective is also related to broader work on depth--width tradeoffs and compute-optimal scaling~\citep{lu2017expressive,cheng2017survey,hoffmann2022training}.

\section{Discussion and Limitations}

Our results suggest that temporal recurrence does not eliminate sequential computation, but changes where it occurs. Non-recurrent models must complete the computation required for each output within the current step, whereas recurrent models can preserve intermediate representations and continue refining them as new inputs arrive. In the settings studied here, this shifts the best observed compute allocation away from within-step depth and toward parallel capacity. Our Sokoban probing results are consistent with this interpretation: the shallower recurrent agent initially forms less complete plans than the deeper agent, but catches up over subsequent environment steps while achieving comparable task performance.

This tradeoff should depend on the structure of the task. Computation can accumulate across time only when consecutive inputs contribute to a shared problem and the carried state preserves useful intermediate computation. We therefore expect recurrence to reduce the need for within-step depth most clearly in settings with persistent context, progressively evolving representations, or opportunities to refine decisions over multiple steps. The same benefit may not arise when each input presents an independent problem, when every decision must be solved immediately, or when the recurrent state is too limited to retain useful intermediate information.

Our empirical study covers two domains, a finite set of architectures, and a restricted range of compute budgets, training durations, and task difficulties. The reported allocations are therefore the best among those evaluated, rather than globally compute-optimal configurations. Isolating the effect of recurrence also requires compromises in architectural parity. ConvLSTM contains state machinery that may favor the recurrent variant, while ConvGLU provides a more direct comparison in which recurrence is introduced through feedback from the previous step's final representation. Similarly, recurrent Transformers require stabilization and memory-management choices that are absent from the standard non-recurrent baseline. Changes in architecture, scale, or architecture-specific tuning may shift the preferred allocation for either model class.

Optimization may also contribute to the earlier depth saturation observed in recurrent models. We train recurrent models using truncated backpropagation through time, which may make credit assignment through deep recurrent updates more difficult~\citep{bengio1994learning,pascanu2013difficulty}. Non-recurrent models constructed from similar blocks continue to benefit from greater depth, arguing against a general inability to optimize deep networks, but this does not rule out optimization difficulties specific to the recurrent setting. Finally, our work proxy captures approximately matched computation rather than realized hardware efficiency. Although reallocating computation from depth to parallel capacity shortens the serial path through each update, actual latency depends on hardware utilization, memory bandwidth, kernel efficiency, and communication overhead. We provide preliminary latency and throughput measurements in Appendix~\ref{sec:appendix-fineweb-hardware-efficiency}, but leave a comprehensive systems evaluation to future work.

\section{Conclusion} 

Across Sokoban and streamed FineWeb, temporal recurrence shifts the best observed compute allocation toward substantially fewer layers per step, with comparable or better performance than non-recurrent models under approximately matched per-step work.
For example, in the small-budget language modeling FineWeb comparison, a recurrent model with $L=2$ reaches a validation loss comparable to the best non-recurrent allocation at $L=16$.
Recurrence and parallel capacity play distinct roles in this shift: recurrence reduces the performance cost of using fewer layers, while per-layer expert count and width provide useful ways to spend the reallocated work.
Increasing parallel capacity remains useful both with and without recurrence.

Our Sokoban decoding analysis is consistent with computation being redistributed across time, with the shallower model catching up in planning while achieving comparable task performance. 
Overall, our results support the broader hypothesis that temporal recurrence can shift the best compute allocation toward fewer layers per step in other streaming tasks where useful computation can accumulate across time.
This motivates exploring shallower, more parallel updates as a potential route to reducing latency and improving training throughput, making general nonlinear recurrent models more practical at scale.

\section*{Acknowledgment}
We acknowledge the support from the Canada CIFAR AI Chair Program and from the Canada Excellence Research Chairs Program. The research was enabled in part by computational resources provided by the Digital Research Alliance of Canada and Mila Quebec AI Institute. IA thanks  Andrei Mircea Romascanu for helpful discussions and comments. 
We thank the Frontier Intelligence Group at Sakana AI for a helpful discussion and suggestions on the paper title, especially Ciaran Regan, whose suggestions helped shape the final title.
We would also like to thank the Python community \citep{van1995python, 4160250} for developing tools that enabled this work, including NumPy \citep{harris2020array}, Matplotlib \citep{hunter2007matplotlib}, Jupyter \citep{2016ppap}, and Pandas \citep{McKinney2013Python}.

\bibliography{references}
\bibliographystyle{iclr2027_conference}

\clearpage
\appendix

\section*{\LARGE \bfseries Appendix Contents}
\addcontentsline{toc}{section}{Appendices}

\startcontents[appendix]
\printcontents[appendix]{l}{1}{\setcounter{tocdepth}{2}}

\clearpage

\section{Architectures}
\label{sec:appendix-architectures}

\begin{figure}[h]
    \centering
    \includegraphics[width=0.6\linewidth]{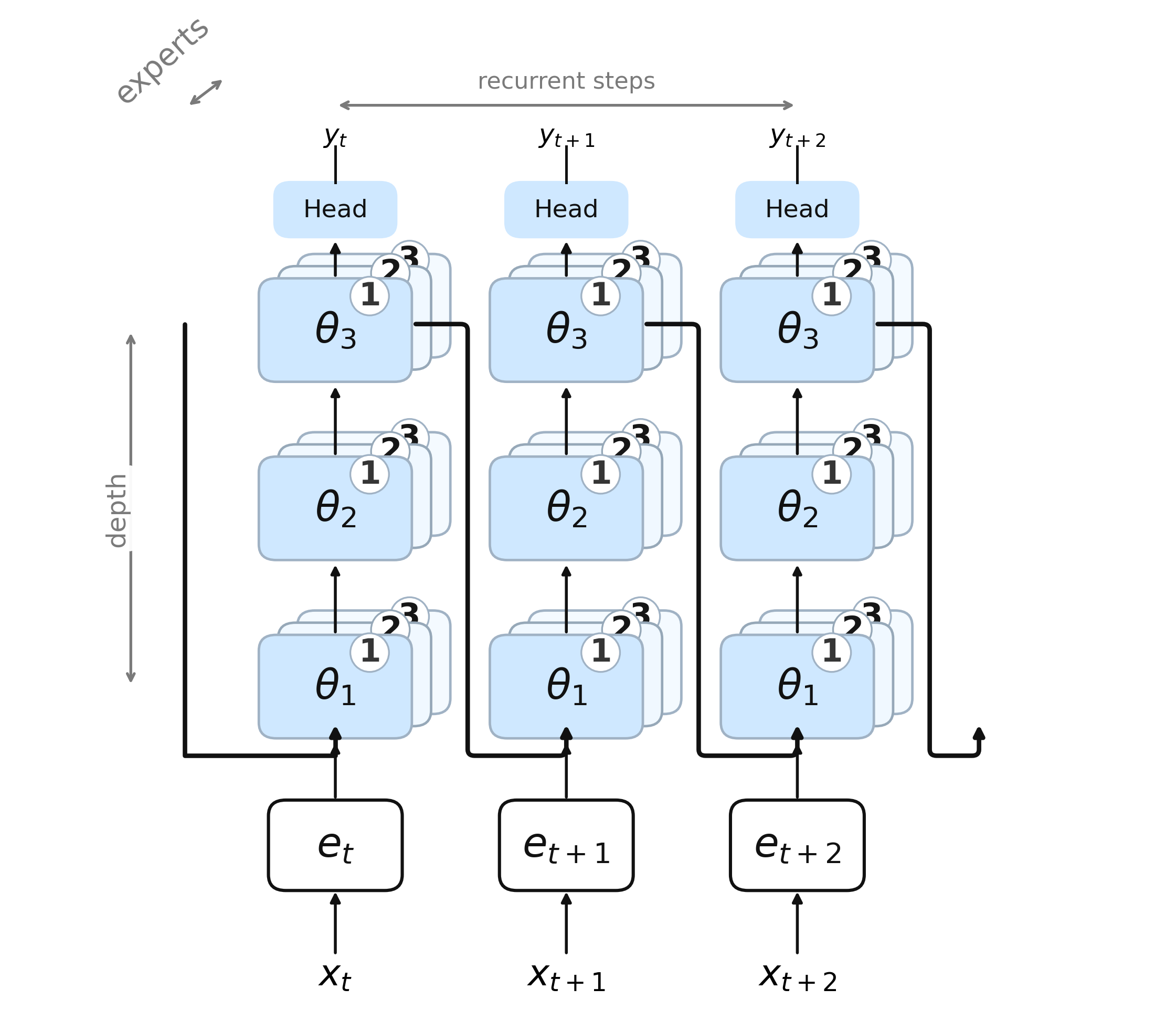}
    \caption{\textbf{Architecture.}
    Each column is one external step. At every step, a fresh input \(x_t\) is encoded, within-step experts update the carried state, a head produces output \(y_t\), and the top state is optionally routed into the next external step in case of recurrence.
    }
    \label{fig:appendix-recurrent-steps-architecture}
\end{figure}

\begin{figure}[htbp]
    \centering
    \includegraphics[width=\linewidth]{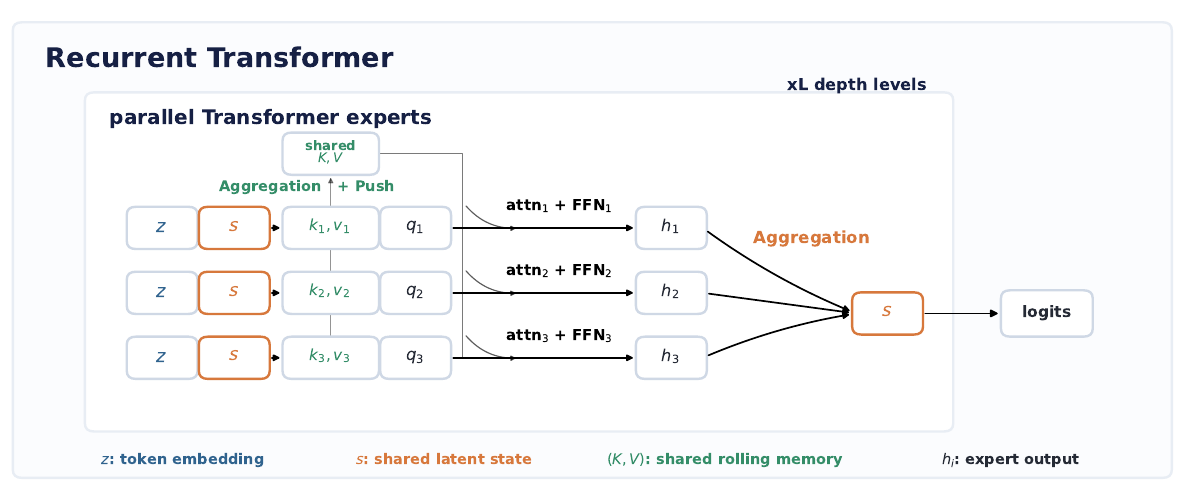}
    \caption{\textbf{Parallel experts Recurrent Transformer.}
    A token embedding \(z\) and shared latent state \(s\) enter each depth level.
    Recurrent Transformer experts run in parallel, propose \((k_i,v_i,q_i)\), aggregate and push \((k_i,v_i)\) into a shared rolling \((K,V)\), then each expert attends to that KV cache, and produces output \(h_i\).
    Expert outputs aggregate into \(s\), which is read by the language-model head.}
    \label{fig:appendix-fineweb-transformer-flow}
\end{figure}

All architecture variants follow the same pattern: an external input is encoded into \(z\), a shared latent state \(s\) is read by several experts in parallel, expert outputs are aggregated, and the updated shared state is carried forward through the external sequence in the recurrent version, as illustrated in Fig.~\ref{fig:appendix-recurrent-steps-architecture}.
The reinforcement-learning models use LSTM experts, while the FineWeb next-token model uses transformer experts.
Figs.~\ref{fig:appendix-rl-pre-flow} and~\ref{fig:appendix-fineweb-transformer-flow} then specialize the expert block for reinforcement learning and FineWeb next-token prediction, and Algorithms~\ref{alg:parallel-recurrent-experts}--\ref{alg:parallel-recurrent-transformer} give the high-level update rules.

\subsection{LSTM}

\begin{figure}[htbp]
    \centering
    \includegraphics[width=\linewidth]{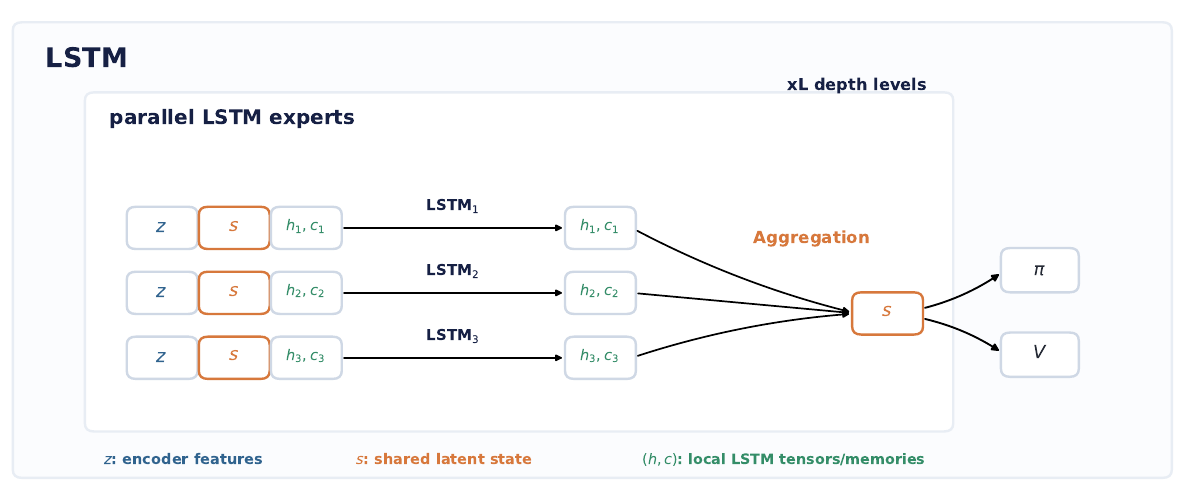}
    \caption{\textbf{Parallel experts LSTM.}
    Each LSTM expert receives encoder features \(z\), the shared latent state \(s\), $h_i$ from the previous depth level or external step and its local recurrent context $c_i$.
    At each depth level, expert outputs are aggregated into the shared latent state \(s\), which is passed to the next depth level. The policy and value heads read \(s\) after the final depth level. }
    \label{fig:appendix-rl-pre-flow}
\end{figure}

Fig.~\ref{fig:appendix-rl-pre-flow} illustrates one depth level with parallel LSTM experts used in reinforcement learning experiments.
Each expert receives the encoded observation \(z\), the shared latent state \(s\), and its own local LSTM tensors \((h_i,c_i)\).
The experts are evaluated in parallel.
At each depth level, their hidden outputs are aggregated into the shared state \(s\), which is passed to the next depth level; the policy and value heads read \(s\) after the final depth level. In the non-recurrent LSTM version, all  $c_i$ are always zero, and $h_i$ and $s$ inputs for the experts in the first depth level are omitted.

\begin{algorithm}[htbp]
\caption{Parallel experts LSTM}
\label{alg:parallel-recurrent-experts}
\begin{algorithmic}[1]
\State Encode external input \(z \gets e(x_t)\)
\State Receive or initialize shared latent state \(s\), rolling hidden states \(h_{1:E}\), and cells \(c_{1:E}^{1:L}\)
\For{within-step depth layer \(\ell=1,\dots,L\)}
    \ForAll{experts \(i=1,\dots,E\) in parallel}
        \State \((h_i, c_i^\ell) \gets f_{\theta_i^\ell}(z, s, h_i, c_i^\ell)\)
    \EndFor
    \State \(s \gets A(h_1,\dots,h_E)\)
\EndFor
\State \((\pi_t,V_t) \gets \mathrm{head}(s)\)
\State Return \(\pi_t\), \(V_t\), updated shared latent state \(s\), and LSTM states \(h_{1:E}, c_{1:E}^{1:L}\)
\end{algorithmic}
\end{algorithm}

Algorithm~\ref{alg:parallel-recurrent-experts} describes the same computation across $L$ sequential depth levels. 
The cell $c_i^\ell$ is local to expert $i$ at depth level $\ell$.

The Sokoban actor--critic use a spatial recurrence. The observation encoder produces a spatial feature map, and the shared latent state keeps the same board-like spatial structure. The encoder is RMS-normalized over the spatial dimensions. At each environment step, recurrent convolutional LSTM (ConvLSTM) experts read the encoded observation, the shared spatial state, and the incoming hidden state and their previous cell states. The ConvLSTM gates are produced by a \(3\times3\) convolution followed by RMS normalization of the gate preactivations.
Expert hidden outputs are aggregated into the shared state, which is carried to the next environment step in case of recurrence or to the next depth level.
The Sokoban head aggregates the spatial expert output with attention pooling and global average pooling, applies a dense GLU, and then uses separate linear projections for action logits and the scalar value estimate.

\subsection{Transformer}

The FineWeb architecture uses standard NanoGPT baseline \citep{vaswani2017attention, Karpathy2022, modded_nanogpt_2024} components: learned token embeddings, grouped-query attention with RoPE, RMS-normalized queries and keys, a squared-ReLU feed-forward block, final RMSNorm, an LM head, and logit soft-capping. To make a parallel expert version of this non-recurrent Transformer we simply execute $E$ transformer blocks in parallel and then merge the outputs using the aggregation in Eq.~ \ref{eq:update}.  

\subsection{Recurrent Transformer}

To make the recurrent version of it we were needed to introduce two changes in addition to recurrence to stabilize the training, and address memory constrains. We have tried to keep these modifications minimal. First, to stabilize the training each block at all depth levels recieves embeddings of the current token in addition to residual stream. Plus, the output of each expert is normilized with parameter-free RMSNorm. The second addition is that we share the KV cache between experts at each depth level. In our experiments, sharing the KV cache increases validation loss by approximately 0.02, but helps greatly to reduce memory pressure and increases throughput.

In detail, in the parallel expert Recurrent Transformer each expert reads the token embedding \(z\) and shared latent state \(s\), forms an expert input, and prepares \(k_i,v_i,q_i\). The per-expert queries remain separate, while the key--value proposals are aggregated across experts into one shared key--value entry.
This entry is pushed into a rolling KV cache \((K,V)\) local to that depth level, so all experts at that level attend to the same shared KV cache.
The rolling KV cache is a fixed-length FIFO buffer: each new aggregated KV entry is appended, and the oldest entry is discarded once the buffer is full.
The expert attention outputs pass through the transformer feed-forward path, and RMS normalization, producing expert outputs \(h_i\) that are aggregated into the next shared latent state.
After the final depth level, the shared state is read by the head.
The KV caches and shared state are then carried forward by the streamed-document language-modeling pipeline.

\begin{algorithm}[htbp]
\caption{Parallel experts Recurrent Transformer}
\label{alg:parallel-recurrent-transformer}
\begin{algorithmic}[1]
\State Encode token \(z \gets e(x_t)\)
\State Receive or initialize shared latent state \(s\) and FIFO KV caches \((K^{1:L},V^{1:L})\)
\For{within-step depth level \(\ell=1,\dots,L\)}
    \ForAll{experts \(i=1,\dots,E\) in parallel}
        \State \((k_i,v_i,q_i) \gets p_{\theta_i^\ell}(z, s)\)
    \EndFor
    \State \(k \gets A(k_1,\dots,k_E)\)
    \State \(v \gets A(v_1,\dots,v_E)\)
    \State \((K^\ell,V^\ell) \gets \mathrm{push\_FIFO}\big((K^\ell,V^\ell), (k,v)\big)\)
    \ForAll{experts \(i=1,\dots,E\) in parallel}
        \State \(h_i \gets g_{\theta_i^\ell}(q_i, K^\ell, V^\ell)\)
    \EndFor
    \State \(s \gets A(h_1,\dots,h_E)\)
\EndFor
\State \(\mathrm{logits}_t \gets \mathrm{head}(s)\)
\State Return next-token logits \(\mathrm{logits}_t\), shared latent state \(s\), and KV caches \((K^{1:L},V^{1:L})\)
\end{algorithmic}
\end{algorithm}

\FloatBarrier

\section{FineWeb Streaming Implementation}
\label{sec:appendix-nanogpt-streaming}

For a shard with $S$ tokens and global batch size $B$, the loader partitions the shard into $B$ contiguous streams of length $\lfloor S/B \rfloor$.
Batch slot $b$ reads only from stream $b$.
At stream position $q$, slot $b$ reads a window of length $T+1$, where $T$ is the training sequence length.
The first $T$ tokens are inputs and the next $T$ shifted tokens are labels.

The recurrent state is also organized by batch slot.
It is reset at the first window of a shard, carried across later adjacent windows in the same slot, and reset again when a new shard begins.

\begin{algorithm}[h]
\caption{FineWeb shard streaming}
\label{alg:nanogpt-streaming}
\begin{algorithmic}[1]
\State Let $B$ be the global batch size and $T$ the training sequence length
\ForAll{training shards with $S$ tokens}
    \State Split the shard into $B$ contiguous streams of length $\lfloor S/B \rfloor$
    \State Initialize recurrent state for all batch slots
    \For{$q = 0, T, 2T, \dots$ while all streams have $T+1$ tokens available}
        \ForAll{batch slots $b=1,\dots,B$}
            \State $w_b \gets$ tokens from stream $b$ at positions $q:q+T+1$
            \State $x_b \gets w_b[0:T]$, \quad $y_b \gets w_b[1:T+1]$
            \State Reset slot state only if $q=0$
        \EndFor
        \State Compute next-token loss and updated recurrent state on $(x_{1:B},y_{1:B})$
        \State Update parameters; carry the detached recurrent state to the next window
    \EndFor
\EndFor
\end{algorithmic}
\end{algorithm}



\section{Sokoban Solve Rate Results}
\label{sec:appendix-convlstm-unfiltered-test}

We evaluate Sokoban policies trained on 50M environment steps on the 1,000-level
\texttt{unfiltered-test} split with ConvLSTM architecture. For each fixed-compute budget and within-step depth $L\in\{1,2,4,8,16\}$, we select the configuration with the highest mean training return. All selected policies act greedily for at most 120 steps per test level.
Table~\ref{tab:appendix-sokoban-convlstm-unfiltered} reports the results.

\begin{table}[h!]
    \centering
    \footnotesize
    \setlength{\tabcolsep}{4pt}
    \renewcommand{\arraystretch}{0.92}
    \caption{\textbf{Sokoban ConvLSTM Unfiltered Test solve rates.} Mean and standard deviation over three seeds are reported. State denotes whether recurrent state is carried or reset at every environment step.}
    \label{tab:appendix-sokoban-convlstm-unfiltered}
    \begin{minipage}[t]{0.49\textwidth}
    \centering
    {\small\strut Recurrent}\\[1pt]
    \begin{tabular}{@{}lrrrc@{}}
        \toprule
        Budget & $L$ & $E$ & $d$ & Unfiltered Test (\%) \\
        \midrule
        small & 1  & 1 & 128 & $86.03 \pm 3.71$ \\
        small & 2  & 2 & 64  & $91.93 \pm 1.08$ \\
        small & 4  & 1 & 64  & $92.70 \pm 1.04$ \\
        small & 8  & 2 & 32  & $86.13 \pm 2.40$ \\
        small & 16 & 1 & 32  & $79.00 \pm 0.96$ \\
        \midrule
        medium & 1  & 1 & 256 & $93.23 \pm 1.06$ \\
        medium & 2  & 2 & 128 & $96.20 \pm 0.70$ \\
        medium & 4  & 1 & 128 & $96.67 \pm 0.67$ \\
        medium & 8  & 2 & 64  & $93.40 \pm 3.84$ \\
        medium & 16 & 1 & 64  & $88.67 \pm 1.72$ \\
        \midrule
        large & 1  & 4  & 512 & $96.23 \pm 0.91$ \\
        large & 2  & 8  & 256 & $98.47 \pm 0.15$ \\
        large & 4  & 1  & 512 & $98.70 \pm 0.17$ \\
        large & 8  & 7  & 128 & $95.67 \pm 1.44$ \\
        large & 16 & 13 & 64  & $88.57 \pm 3.80$ \\
        \bottomrule
    \end{tabular}
    \end{minipage}
    \hfill
    \begin{minipage}[t]{0.49\textwidth}
    \centering
    {\small\strut Feedforward}\\[1pt]
    \begin{tabular}{@{}lrrrc@{}}
        \toprule
        Budget & $L$ & $E$ & $d$ & Unfiltered Test (\%) \\
        \midrule
        small & 1  & 4  & 64  & $3.20 \pm 0.00$ \\
        small & 2  & 2  & 64  & $30.50 \pm 0.96$ \\
        small & 4  & 1  & 64  & $59.63 \pm 3.83$ \\
        small & 8  & 2  & 32  & $74.63 \pm 3.91$ \\
        small & 16 & 1  & 32  & $73.23 \pm 2.97$ \\
        \midrule
        medium & 1  & 4   & 128 & $3.93 \pm 0.12$ \\
        medium & 2  & 2   & 128 & $39.03 \pm 0.78$ \\
        medium & 4  & 1   & 128 & $68.93 \pm 1.95$ \\
        medium & 8  & 2   & 64  & $86.73 \pm 1.34$ \\
        medium & 16 & 1   & 64  & $84.10 \pm 4.32$ \\
        \midrule
        large & 1  & 58 & 128 & $4.57 \pm 0.50$ \\
        large & 2  & 2  & 512 & $52.93 \pm 0.68$ \\
        large & 4  & 4  & 256 & $78.47 \pm 0.55$ \\
        large & 8  & 7  & 128 & $92.00 \pm 0.26$ \\
        large & 16 & 13 & 64  & $93.00 \pm 0.40$ \\
        \bottomrule
    \end{tabular}
    \end{minipage}
\end{table}
\FloatBarrier

For external context, \citet{chung2023thinker} report Thinker and a reproduction of DRC~\citep{guez2019investigation} on the same unfiltered Sokoban test set after 50 million real-environment frames. \citet{taufeeque2024planning} report DRC(3,3) and ResNet point estimates after two billion environment steps on the same split with the same 120-step episode limit.

\begin{table}[H]
    \centering
    \small
    \caption{\textbf{Reported Sokoban baselines on Unfiltered Test.}
    The 50M-frame entries are the mean with one standard deviation reported by
    \citet{chung2023thinker}; the 2B-step entries are reported
    by \citet{taufeeque2024planning}.}
    \label{tab:appendix-sokoban-literature-baselines}
    \begin{tabular}{@{}lcc@{}}
        \toprule
        Method & Training steps & Unfiltered Test \\
        \midrule
        Thinker & 50M & $94.5 \pm 0.8$ \\
        DRC(3,3) (Thinker reproduction) & 50M & $91.3 \pm 1.2$ \\
        DRC(3,3) & 2B & $99.3$ \\
        ResNet & 2B & $97.9$ \\
        \bottomrule
    \end{tabular}
\end{table}

\FloatBarrier

\section{Aggregation Ablations}
\label{sec:appendix-aggregation-ablation}

\begin{figure}[h]
    \centering
    \includegraphics[width=0.98\linewidth]{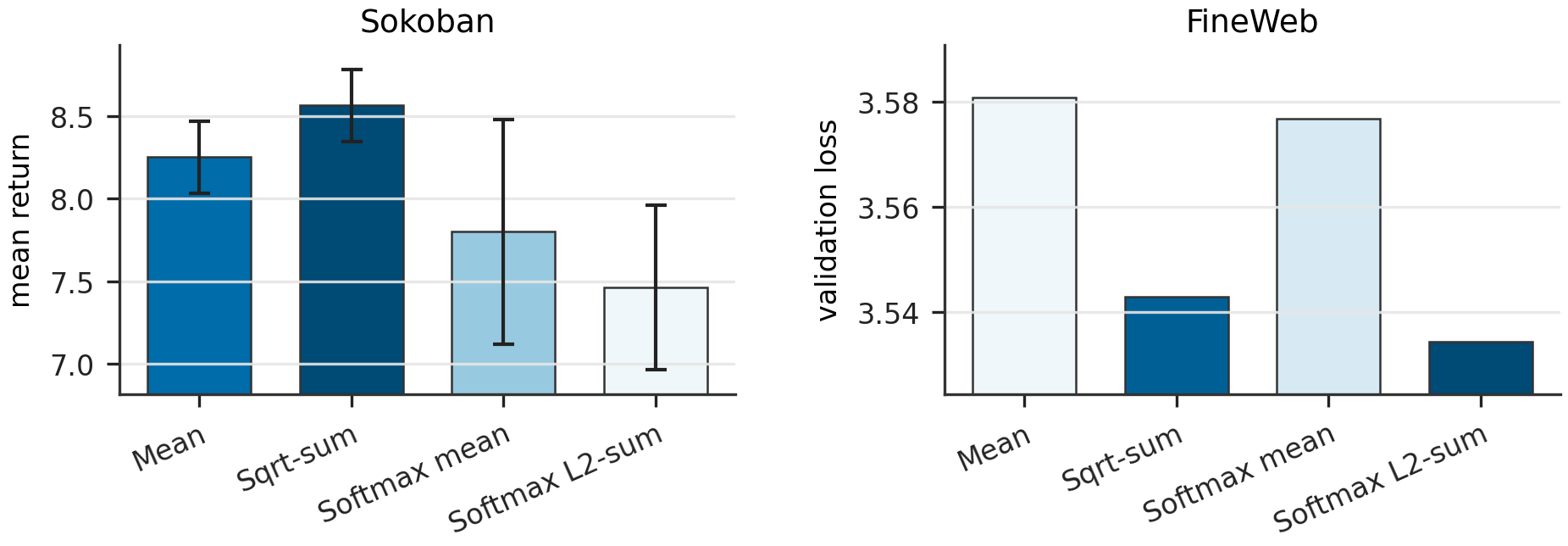}
    \caption{\textbf{Aggregation ablations across domains.} \textbf{Left}: Sokoban ConvLSTM with carried state with \(d=128,L=1,E=4\); bars show mean episodic return with one standard error across three seeds. \textbf{Right}: FineWeb next-token prediction, lower is better. }
    \label{fig:appendix-aggregation-ablations}
\end{figure}

\paragraph{Aggregation rules.}
Let \(h_i\) denote the output of expert \(i\). 
For routed variants, \(\alpha_i\) denotes the learned softmax routing coefficient for expert \(i\).
The four ablations choices are mean pooling, sqrt-normalized sum pooling, softmax-gated weighted averaging, and L2-normalized softmax-gated aggregation.
We label these as Mean, Sqrt-sum, Softmax mean, and Softmax L2-sum in Fig.~\ref{fig:appendix-aggregation-ablations}:
\begin{align}
    y_g^{\mathrm{mean}}
    &= \frac{1}{m} \sum_{i\in \mathcal{E}_g} h_i,
    \\
    y_g^{\sqrt{\mathrm{sum}}}
    &= \frac{1}{\sqrt{m}} \sum_{i\in \mathcal{E}_g} h_i,
    \\
    y_g^{\mathrm{rmean}}
    &= \frac{\sum_{i\in \mathcal{E}_g} \alpha_i h_i}{\sum_{i\in \mathcal{E}_g} \alpha_i},
    \\
    y_g^{\mathrm{r}\sqrt{\mathrm{sum}}}
    &= \frac{\sum_{i\in \mathcal{E}_g} \alpha_i h_i}{\sqrt{\sum_{i\in \mathcal{E}_g} \alpha_i^2}}.
\end{align}
The sqrt-normalized variants are variance-preserving normalizations: if independent expert outputs have comparable scale, the aggregate scale stays roughly stable as the effective number of contributors changes.
The routed mean is the usual softmax-gated weighted average.
In the implementation, denominators in the routed formulas are protected with a small \(\epsilon\) for numerical stability.

\paragraph{Routing coefficients.}
The coefficients are produced by a scaled dot-product router.
For dense experts, the router context is the concatenation of the encoded observation or token representation and the current shared latent state,
\(c=[x_{\mathrm{enc}}; s]\).
For convolutional recurrent experts, both tensors are first spatially average-pooled, then concatenated.
The router maps this context to a query and compares it to learned expert keys:
\begin{align}
    q &= W_q c + b_q,
    &
    k_i &= W_k e_i + b_k,
    &
    \ell_i &= \frac{q^\top k_i}{\sqrt{d}},
    &
    \alpha_i &= \operatorname{softmax}(\ell)_i.
\end{align}
Here \(e_i\) is a learned embedding for expert \(i\), \(d\) is the router embedding dimension, and \(\ell_i\) are the router logits.

\paragraph{Default choice.}
We use Sqrt-sum as the default aggregation rule used throughout the paper, except in experiments that explicitly ablate aggregation.
It is parameter-free, keeps the aggregate scale stable as the number of contributing experts changes, and is consistently competitive across domains.
For the FineWeb shared-global transformer, we use it for both expert outputs and the merged key-value proposals.

\FloatBarrier

\section{Communication Radius}
\label{sec:appendix-communication-radius}

We vary how many other experts each expert can read at each depth level in a Sokoban model with \(E=8\) ConvLSTM experts per layer, width \(d=32\), and within-step depth $L=2$.
Experts are arranged in a ring and read their own state plus those of a chosen number of preceding neighbours.
The visible states are summed and divided by the square root of their count, keeping the input width and expert computation fixed.

\begin{figure}[htbp]
    \centering
    \includegraphics[width=0.6\linewidth]{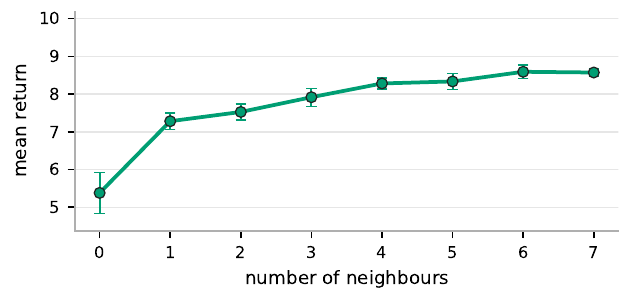}
    \caption{\textbf{Communication between experts in Sokoban.}
    Mean episodic return versus the number of neighbors. Mean returns with one standard error are reported across three seeds.}
    \label{fig:appendix-sokoban-communication-topology}
\end{figure}

Fig.~\ref{fig:appendix-sokoban-communication-topology} shows that full communication achieves the highest mean return. 
These results indicate that sharing information across experts contributes to the model's performance. 
We  only use full communication for other experiments in the paper.

\FloatBarrier
\clearpage
\section{Sokoban Planning Analysis}
\label{sec:appendix-sokoban-agent-move-decoding}

We follow the spatial probing method of \citet{bush2025interpreting}, who decoded iteratively refined plans in DRC agents. We apply their agent-move probes to our models' final residual representations and compare how decoded plans develop across environment steps at different depth allocations.

We extend Fig.~\ref{fig:sokoban-agent-move-recall} to four large-budget recurrent ConvLSTMs with fixed width $d=256$ and $(L,E) \in (1,15),(2,8),(4,4),(8,2)$. For each allocation, we select the best of three training seeds by solve rate on 1,000 Unfiltered Test levels, using greedy actions with a 120-step limit.

Following \citet{bush2025interpreting} for each frozen policy, we train  convolution $3\times3$ probes on 3,000 Unfiltered Train trajectories.
Each probe reads the final residual representation at each step.
At each level location, the target is the direction of the agent in this location in the recorded trajectory, or \textsc{Never} if agent does not visit it. 

We evaluate decoding on the 906 Test levels solved by all four
policies. Recall is the fraction of non-\textsc{Never} targets correctly predicted above threshold, and precision is the fraction of emitted directions that are correct.
We choose one fixed threshold per model to achieve approximately 93\% precision for all prediction over first $20$ steps.

\begin{figure}[H]
    \centering
    \includegraphics[width=\linewidth]{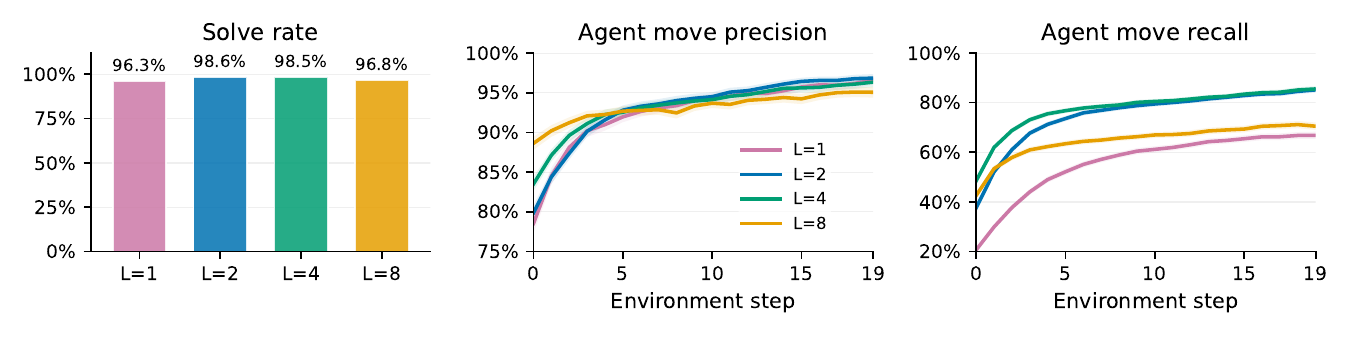}
    \caption{Agent-move decoding comparison across models with different $L$.}  
    \label{fig:appendix-sokoban-agent-move-depths}
\end{figure}

The $L=4$ model initially has higher recall than the $L=2$ model, indicating a more complete decoded plan. The $L=2$ model catches up over the next few environment steps, after which both maintain similar recall. Together with their nearly identical solve rates, this pattern is consistent with the shallower model compensating for less planning within each update by continuing the computation across environment steps, catching up early enough to preserve task performance.
A similar pattern appears for the $L=1$ and $L=8$ models, which also achieve similar test solve rates: $L=8$ initially has higher recall, while $L=1$ narrows the gap over subsequent steps, although it does not fully close it.

\begin{figure}[H]
    \centering
    \includegraphics[width=0.85\linewidth]{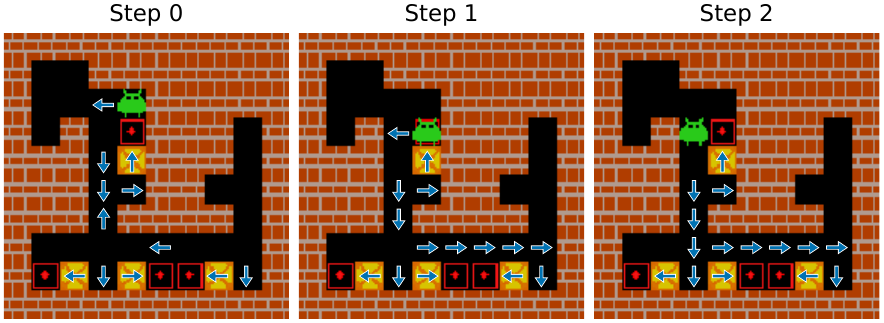}
    \caption{\textbf{Agent move plans over the first three environment steps.} The two layer model from Fig.~\ref{fig:appendix-sokoban-agent-move-depths} is shown on one Test level.}
    \label{fig:appendix-sokoban-agent-move-steps}
\end{figure}

\paragraph{Visualization.}
To make the analysis self-contained, we also include a visualization of how plans develop in the two layer model (Fig.~\ref{fig:appendix-sokoban-agent-move-steps}).
The main DRC model analyzed by \citet{bush2025interpreting} combines within-step looping with temporal recurrence, and our models do not use within-step looping.

\section{Implementation Details and Hyperparameters Used in Experiments}
\label{sec:appendix-hyperparameters}

This section reports the training hyperparameters used across the main experiments.

\begin{table}[htbp]
    \centering
    \caption[Hyperparameters used in reinforcement-learning experiments]{Hyperparameters used in reinforcement-learning experiments.}
    \label{tab:appendix-rl-hyperparameters}
    \begin{tabular}{ll}
        \toprule[0.1ex]
        \textbf{Parameter} & \textbf{Value} \\
        \\
        \textbf{Sokoban} \\
        Encoder hidden dimension & 32 \\
        Discount rate \(\gamma\) & 0.97 \\
        Lambda for general advantage estimation & 0.97 \\
        Entropy coefficient & 0.01 \\
        Value function coefficient & 1.0 \\
        Normalize advantages & No \\
        Unroll length  & 20 \\
        Number of environments & 32 \\
        Total environment steps & 50,000,000 \\
        Learning rate & \(4 \times 10^{-4}\) \\
        Anneal learning rate & Yes \\
        Optimizer & Adam \\
        Adam epsilon & \(10^{-6}\) \\
        Maximum gradient norm for clipping & 1.0 \\
        \bottomrule[0.25ex]
    \end{tabular}
\end{table}

\begin{table}[htbp]
    \centering
    \caption[Hyperparameters used in next-token prediction experiments]{Hyperparameters used in next-token prediction experiments.}
    \label{tab:appendix-nanogpt-hyperparameters}
    \begin{tabular}{ll}
        \toprule[0.1ex]
        \textbf{Parameter} & \textbf{Value} \\
        \\
        \textbf{FineWeb next-token prediction} \\
        Vocabulary size & 50,304 \\
        Attention heads & 8 query, 4 key-value default; scaled with width proportionally \\
        
        Token batch size & 131,072 \\
        Training steps & 10,000 \\
        Optimizer & Muon core; AdamW embed/readout \\
        Gate parameters & Muon \\
        Optimizer betas & (0.8, 0.95) \\
        Gradient clipping & 1.0 \\
        Learning rate schedule & Warmup-cosine decay \\
        Peak learning rate & \(2 \times 10^{-2}\) \\
        Minimum learning rate & \(2 \times 10^{-4}\) \\
        Warmup steps & 100 \\
        Embedding learning rate & \(0.2(d/768)^{-1/2}\) \\
        Readout learning rate & \(0.004(d/768)^{-1/2}\) \\
        Weight decay & 0 \\
        Cautious weight decay & 0.01 \\
        Numerical precision & bfloat16 \\
        \\
        \textbf{Transformer} \\
        Training sequence length & 256 or 2048 \\
        \\
        \textbf{Recurrent  Transformer} \\
        Shared KV cache length & 128 or 1024\\
        Training sequence length & 64 \\
        
    \end{tabular}
\end{table}

\FloatBarrier

\subsection{Compute Resources}
\label{sec:appendix-compute-resources}

The reinforcement-learning experiments were run on NVIDIA RTX 8000 and L40S GPUs. Most Sokoban runs use one or two GPUs and take a few hours for small configurations, while the largest fixed-compute and within-step-depth-heavy configurations can take up to roughly one day. 
The FineWeb next-token prediction experiments were run on NVIDIA H100 GPUs. All FineWeb runs use four GPUs and take several hours for standard configurations, while the largest fixed-compute and within-step-depth-heavy configurations can take up to roughly one day.

\section{Latency and Training Throughput}
\label{sec:appendix-fineweb-hardware-efficiency}

These measurements are preliminary.  Neither implementation was extensively optimized for inference, so the latency results should be interpreted with caution.  Both implementations use JAX; Recurrent Transformer experts at the same depth level are evaluated with \texttt{jax.vmap} whenever they can execute in parallel.  Measurements use four 80GB NVIDIA H100 GPUs, hidden
dimension 768, eight query heads, four key-value heads, and budget \(L \cdot E=16\).  For the Recurrent Transformer, \(L\) is within-step depth and \(E\) is the number of parallel experts per depth level.  In latency measurements, \(B\) is the global inference batch.  In training throughput measurements, $B$ is the global batch size and $C$ is the training sequence length, giving $CB$ tokens per optimizer update.

\subsection{Inference Latency}
\label{sec:appendix-latency}

Table~\ref{tab:appendix-fineweb-4gpu-latency} reports latency after pre-filling the KV caches of either model to length \(C\).

\begin{table*}[h]
    \centering
    \scriptsize
    \caption{Four-H100 median latency in milliseconds per
    token. }
    \label{tab:appendix-fineweb-4gpu-latency}
    \begin{tabular}{lcc}
        \toprule
        Model & \(C=128,B=16\) & \(C=2048,B=16\) \\
        \midrule
        Transformer \(L=16,E=1\) & 0.565 & 0.670 \\
        Recurrent \(L=1,E=16\) & 0.108 & 0.131 \\
        Recurrent \(L=2,E=8\) & 0.431 & 0.554 \\
        Recurrent \(L=4,E=4\) & 0.552 & 0.637 \\
        Recurrent \(L=8,E=2\) & 0.815 & 0.935 \\
        Recurrent \(L=16,E=1\) & 1.402 & 1.577 \\
        \bottomrule
    \end{tabular}
\end{table*}

\subsection{Training Throughput}

Table~\ref{tab:appendix-fineweb-4gpu-training-throughput} measures the complete
FineWeb training update, including the normal dataloader, forward and backward
passes, and optimizer update.

\begin{table*}[h]
    \centering
    \scriptsize
    \setlength{\tabcolsep}{3.2pt}
    \caption{Four-H100 FineWeb training throughput in tokens/s.
    Each panel has fixed number of tokens per optimizer update.\(C\) is the training sequence length  and  \(B\) is the global batch size.}
    \label{tab:appendix-fineweb-4gpu-training-throughput}
    \begin{tabular}{lcccccc}
        \toprule
        \multicolumn{7}{c}{131k effective tokens per optimizer update} \\
        \cmidrule(lr){1-7}
        Model & \shortstack{\(C=32\)\\\(B=4{,}096\)} & \shortstack{\(C=64\)\\\(B=2{,}048\)} & \shortstack{\(C=128\)\\\(B=1{,}024\)} & \shortstack{\(C=256\)\\\(B=512\)} & \shortstack{\(C=1024\)\\\(B=128\)} & \shortstack{\(C=2048\)\\\(B=64\)} \\
        \midrule
        Transformer \(L=16,E=1\) & 967.7k & 1{,}027.4k & 1{,}025.3k & 1{,}054.3k & 1{,}016.9k & 968.1k \\
        Recurrent \(L=1,E=16\) & 505.4k & 445.7k & 349.5k & 229.4k & 51.1k & OOM \\
        Recurrent \(L=2,E=8\) & 265.8k & 209.5k & 145.2k & 85.8k & OOM & OOM \\
        Recurrent \(L=4,E=4\) & 189.7k & 148.7k & 99.1k & OOM & OOM & OOM \\
        Recurrent \(L=8,E=2\) & 118.3k & 88.4k & OOM & OOM & OOM & OOM \\
        \bottomrule
    \end{tabular}

    \vspace{0.7em}

    \begin{tabular}{lcccccc}
        \toprule
        \multicolumn{7}{c}{262k effective tokens per optimizer update} \\
        \cmidrule(lr){1-7}
        Model & \shortstack{\(C=32\)\\\(B=8{,}192\)} & \shortstack{\(C=64\)\\\(B=4{,}096\)} & \shortstack{\(C=128\)\\\(B=2{,}048\)} & \shortstack{\(C=256\)\\\(B=1{,}024\)} & \shortstack{\(C=1024\)\\\(B=256\)} & \shortstack{\(C=2048\)\\\(B=128\)} \\
        \midrule
        Transformer \(L=16,E=1\) & 886.9k & 1{,}031.5k & 1{,}095.8k & 1{,}132.0k & 1{,}095.8k & 1{,}071.5k \\
        Recurrent \(L=1,E=16\) & 556.2k & 526.4k & 447.7k & 313.1k & OOM & OOM \\
        Recurrent \(L=2,E=8\) & 306.0k & 264.6k & 190.2k & OOM & OOM & OOM \\
        Recurrent \(L=4,E=4\) & 199.9k & 170.8k & OOM & OOM & OOM & OOM \\
        Recurrent \(L=8,E=2\) & OOM & OOM & OOM & OOM & OOM & OOM \\
        \bottomrule
    \end{tabular}

\end{table*}

\FloatBarrier

\FloatBarrier

\section{FineWeb Fixed-Compute Sweeps }
\label{sec:fineweb-fixed-compute-budgets}

\begin{figure}[h]
    \centering
    \includegraphics[width=0.92\linewidth]{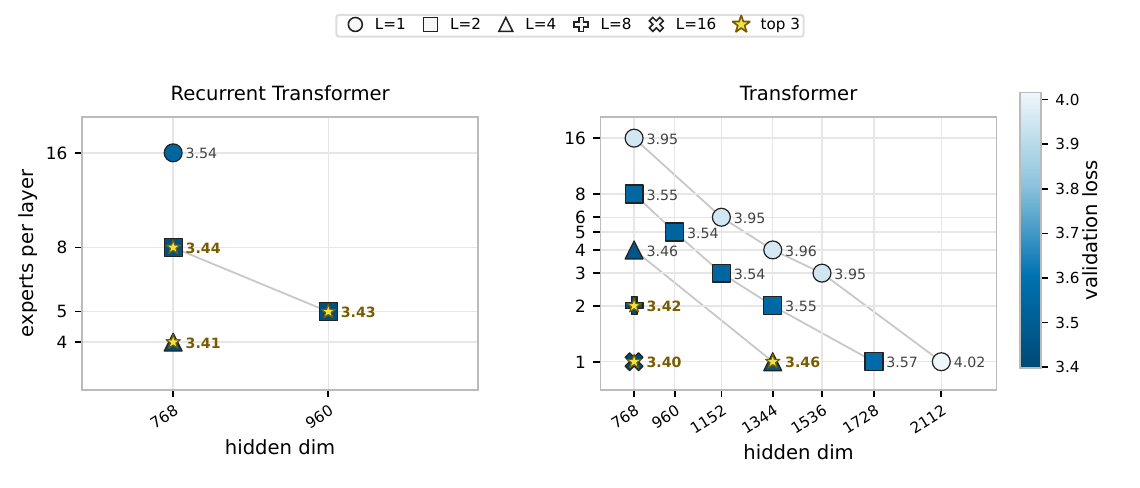}
    \caption{FineWeb small budget allocations with long context. Lower is better.}
    \label{fig:fineweb-recmoe-gptmoe-2048}
\end{figure}

\begin{figure}[h]
    \centering
    \includegraphics[width=0.92\linewidth]{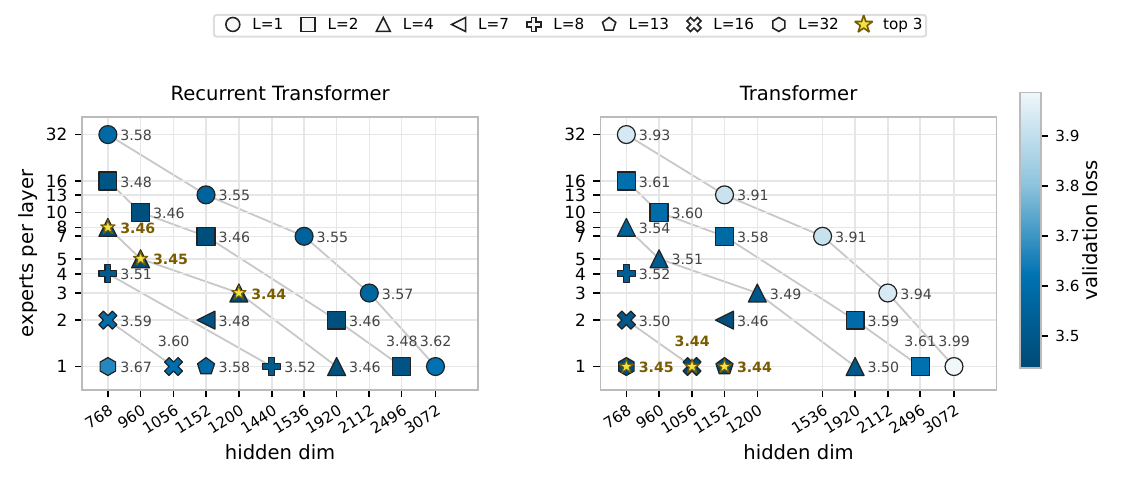}
    \caption{FineWeb medium budget allocations. Lower is better.}
    \label{fig:fineweb-budget32-forward-matched}
\end{figure}

Fig.~\ref{fig:fineweb-recmoe-gptmoe-2048} show FineWeb fixed-compute allocations at small budget with long context and Fig.~\ref{fig:fineweb-budget32-forward-matched} shows FineWeb fixed-compute allocations the medium budget, comparing Recurrent Transformer and Transformer architectures.

\FloatBarrier


\section{Sokoban Fixed-Compute Sweeps}
\label{sec:appendix-convlstm-fixed-compute-sweeps}

\begin{figure*}[h!]
    \centering
    \includegraphics[width=0.98\textwidth]{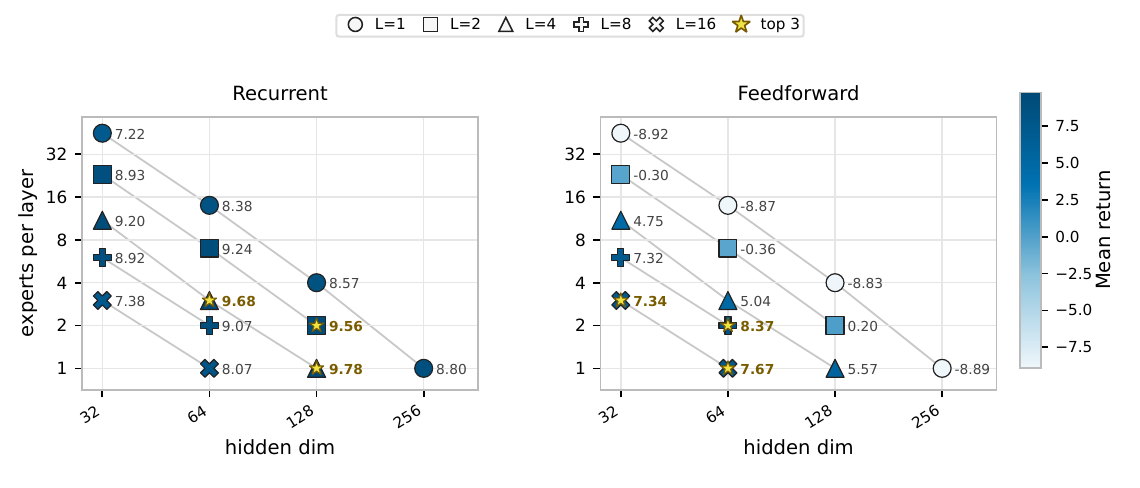}
    \caption{Sokoban medium budget allocations.}
    \label{fig:appendix-sokoban-medium-lstm-recurrent-feedforward}
\end{figure*}

\begin{figure*}[h!]
    \centering
    \includegraphics[width=0.98\textwidth]{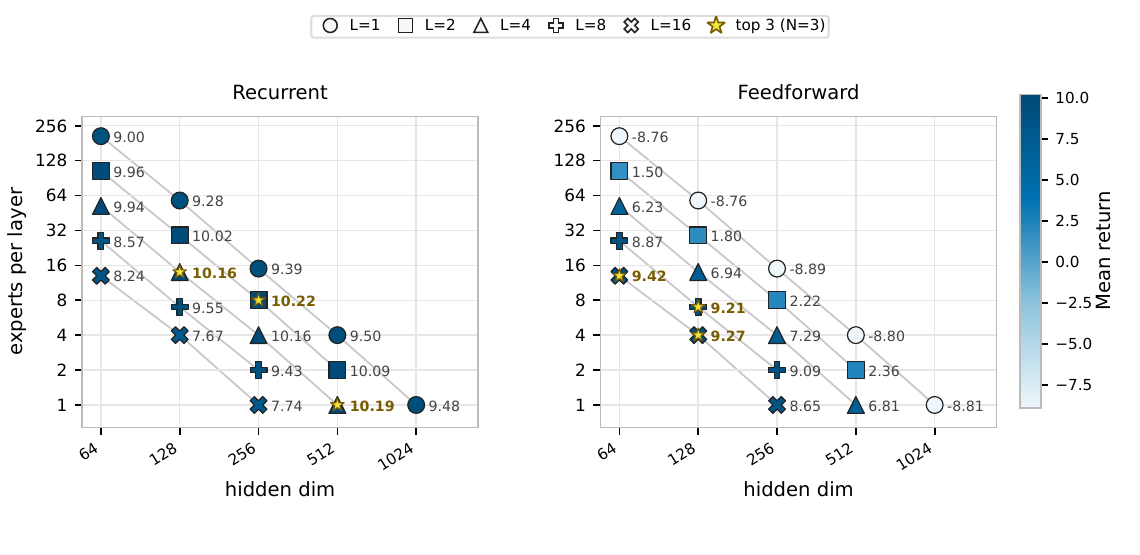}
    \caption{Sokoban large budget allocations.}
    \label{fig:appendix-sokoban-large-lstm-recurrent-feedforward}
\end{figure*}

Fig.~\ref{fig:appendix-sokoban-small-lstm-recurrent-feedforward}, Fig.~\ref{fig:appendix-sokoban-medium-lstm-recurrent-feedforward} and Fig.~\ref{fig:appendix-sokoban-large-lstm-recurrent-feedforward} show the Sokoban the ConvLSTM allocations at the small, medium, and large budgets under recurrent and feedforward settings. The large-budget snapshot includes only completed training runs.


\begin{table}[H]
\centering
\caption{\textbf{Sokoban ConvLSTM matched-compute budgets run summaries.} Mean, sample variance (denominator \(N-1\)), and seed count \(N\) of the per-seed final-five episodic returns.}
\label{tab:appendix-convlstm-fixed-compute-run-summaries}
\begin{minipage}[t]{0.49\textwidth}
\centering
{\small Recurrent}\\[1pt]
\fontsize{6.0}{6.5}\selectfont
\setlength{\tabcolsep}{2.0pt}
\renewcommand{\arraystretch}{0.92}
\begin{tabular}{@{}lrrrrrr@{}}
\toprule
Budget & \(L\) & \(E\) & \(d\) & Mean & Variance & \(N\) \\
\midrule
Small & 1 & 1 & 128 & 7.750 & 0.0076 & 3 \\
Small & 1 & 4 & 64 & 7.664 & 0.1200 & 3 \\
Small & 1 & 12 & 32 & 5.670 & 1.0279 & 3 \\
Small & 2 & 2 & 64 & 8.947 & 0.0685 & 3 \\
Small & 2 & 6 & 32 & 7.336 & 0.0454 & 3 \\
Small & 4 & 1 & 64 & 9.212 & 0.0005 & 3 \\
Small & 4 & 3 & 32 & 7.952 & 0.4108 & 3 \\
Small & 8 & 2 & 32 & 7.929 & 0.1292 & 3 \\
Small & 16 & 1 & 32 & 6.297 & 0.0277 & 3 \\
\midrule
Medium & 1 & 1 & 256 & 8.795 & 0.0179 & 3 \\
Medium & 1 & 4 & 128 & 8.566 & 0.1426 & 3 \\
Medium & 1 & 14 & 64 & 8.382 & 0.1108 & 3 \\
Medium & 1 & 45 & 32 & 7.224 & 0.3117 & 3 \\
Medium & 2 & 2 & 128 & 9.558 & 0.0101 & 3 \\
Medium & 2 & 7 & 64 & 9.241 & 0.0038 & 3 \\
Medium & 2 & 23 & 32 & 8.928 & 0.0031 & 3 \\
Medium & 4 & 1 & 128 & 9.778 & 0.0197 & 3 \\
Medium & 4 & 3 & 64 & 9.677 & 0.0443 & 3 \\
Medium & 4 & 11 & 32 & 9.200 & 0.0221 & 3 \\
Medium & 8 & 2 & 64 & 9.067 & 0.9128 & 3 \\
Medium & 8 & 6 & 32 & 8.919 & 0.1909 & 3 \\
Medium & 16 & 1 & 64 & 8.068 & 0.0537 & 3 \\
Medium & 16 & 3 & 32 & 7.378 & 0.1338 & 3 \\
\midrule
Large & 1 & 1 & 1024 & 9.479 & 0.0288 & 3 \\
Large & 1 & 4 & 512 & 9.503 & 0.0033 & 3 \\
Large & 1 & 15 & 256 & 9.394 & 0.0132 & 3 \\
Large & 1 & 58 & 128 & 9.278 & 0.0469 & 3 \\
Large & 1 & 208 & 64 & 9.001 & 0.0354 & 3 \\
Large & 2 & 2 & 512 & 10.092 & 0.0085 & 3 \\
Large & 2 & 8 & 256 & 10.219 & 0.0113 & 3 \\
Large & 2 & 29 & 128 & 10.023 & 0.0074 & 3 \\
Large & 2 & 104 & 64 & 9.963 & 0.0267 & 3 \\
Large & 4 & 1 & 512 & 10.189 & 0.0077 & 3 \\
Large & 4 & 4 & 256 & 10.163 & 0.0138 & 3 \\
Large & 4 & 14 & 128 & 10.164 & 0.0074 & 3 \\
Large & 4 & 52 & 64 & 9.944 & 0.0239 & 3 \\
Large & 8 & 2 & 256 & 9.429 & 0.4706 & 3 \\
Large & 8 & 7 & 128 & 9.551 & 0.1403 & 3 \\
Large & 8 & 26 & 64 & 8.575 & 0.2624 & 3 \\
Large & 16 & 1 & 256 & 7.735 & 0.1215 & 3 \\
Large & 16 & 4 & 128 & 7.667 & 0.2837 & 3 \\
Large & 16 & 13 & 64 & 8.241 & 0.7027 & 3 \\
\bottomrule
\end{tabular}
\end{minipage}
\hfill
\begin{minipage}[t]{0.49\textwidth}
\centering
{\small Feedforward}\\[1pt]
\fontsize{6.0}{6.5}\selectfont
\setlength{\tabcolsep}{2.0pt}
\renewcommand{\arraystretch}{0.92}
\begin{tabular}{@{}lrrrrrr@{}}
\toprule
Budget & \(L\) & \(E\) & \(d\) & Mean & Variance & \(N\) \\
\midrule
Small & 1 & 1 & 128 & -9.080 & 0.0007 & 3 \\
Small & 1 & 4 & 64 & -8.940 & 0.0185 & 3 \\
Small & 1 & 12 & 32 & -9.174 & 0.0033 & 3 \\
Small & 2 & 2 & 64 & -2.050 & 0.2530 & 3 \\
Small & 2 & 6 & 32 & -2.255 & 0.2295 & 3 \\
Small & 4 & 1 & 64 & 3.697 & 0.1589 & 3 \\
Small & 4 & 3 & 32 & 2.938 & 0.0237 & 3 \\
Small & 8 & 2 & 32 & 5.902 & 0.5632 & 3 \\
Small & 16 & 1 & 32 & 5.491 & 0.0821 & 3 \\
\midrule
Medium & 1 & 1 & 256 & -8.889 & 0.0042 & 3 \\
Medium & 1 & 4 & 128 & -8.829 & 0.0086 & 3 \\
Medium & 1 & 14 & 64 & -8.874 & 0.0006 & 3 \\
Medium & 1 & 45 & 32 & -8.917 & 0.0004 & 3 \\
Medium & 2 & 2 & 128 & 0.204 & 0.1498 & 3 \\
Medium & 2 & 7 & 64 & -0.357 & 0.2066 & 3 \\
Medium & 2 & 23 & 32 & -0.297 & 0.1262 & 3 \\
Medium & 4 & 1 & 128 & 5.571 & 0.1257 & 3 \\
Medium & 4 & 3 & 64 & 5.039 & 0.1118 & 3 \\
Medium & 4 & 11 & 32 & 4.753 & 0.0918 & 3 \\
Medium & 8 & 2 & 64 & 8.367 & 0.0050 & 3 \\
Medium & 8 & 6 & 32 & 7.324 & 0.0180 & 3 \\
Medium & 16 & 1 & 64 & 7.674 & 0.1781 & 3 \\
Medium & 16 & 3 & 32 & 7.336 & 0.0311 & 3 \\
\midrule
Large & 1 & 1 & 1024 & -8.810 & 0.0143 & 3 \\
Large & 1 & 4 & 512 & -8.801 & 0.0122 & 3 \\
Large & 1 & 15 & 256 & -8.885 & 0.0265 & 3 \\
Large & 1 & 58 & 128 & -8.760 & 0.0083 & 3 \\
Large & 1 & 208 & 64 & -8.765 & 0.0041 & 3 \\
Large & 2 & 2 & 512 & 2.356 & 0.0064 & 3 \\
Large & 2 & 8 & 256 & 2.216 & 0.0352 & 3 \\
Large & 2 & 29 & 128 & 1.797 & 0.0835 & 3 \\
Large & 2 & 104 & 64 & 1.497 & 0.0736 & 3 \\
Large & 4 & 1 & 512 & 6.807 & 0.0178 & 3 \\
Large & 4 & 4 & 256 & 7.295 & 0.0715 & 3 \\
Large & 4 & 14 & 128 & 6.938 & 0.0739 & 3 \\
Large & 4 & 52 & 64 & 6.226 & 0.2677 & 3 \\
Large & 8 & 2 & 256 & 9.090 & 0.0501 & 3 \\
Large & 8 & 7 & 128 & 9.215 & 0.0544 & 3 \\
Large & 8 & 26 & 64 & 8.875 & 0.1905 & 3 \\
Large & 16 & 1 & 256 & 8.650 & 0.1802 & 3 \\
Large & 16 & 4 & 128 & 9.271 & 0.0152 & 3 \\
Large & 16 & 13 & 64 & 9.417 & 0.0358 & 3 \\
\bottomrule
\end{tabular}
\end{minipage}
\end{table}

\section{ConvGLU Sokoban}
\label{sec:appendix-convglu-sokoban}

\begin{figure*}[t]
    \centering
    \begin{minipage}[c]{0.48\textwidth}
        \centering
        \includegraphics[width=\linewidth]{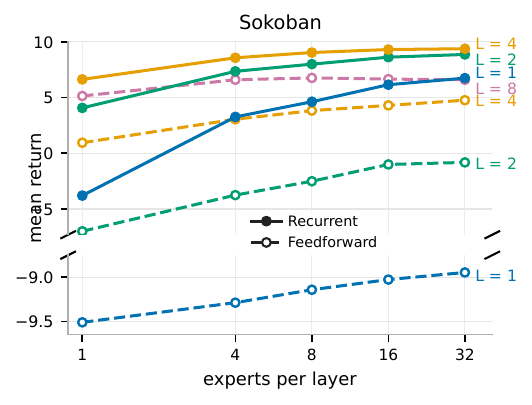}
    \end{minipage}\hfill
    \begin{minipage}[c]{0.48\textwidth}
        \centering
        \includegraphics[width=\linewidth]{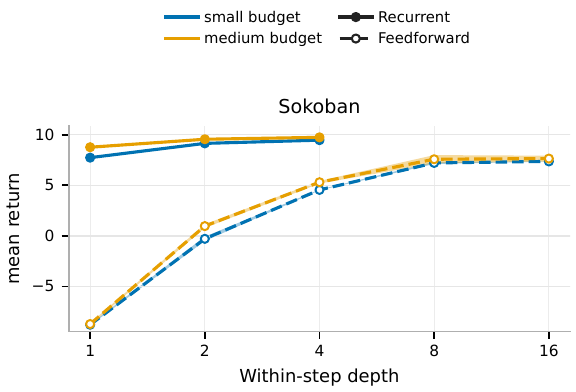}
    \end{minipage}
    \caption{\textbf{Sokoban ConvGLU scaling.} \textbf{Left:} Mean return across different numbers of experts per layer at fixed within-step depths. \textbf{Right:} Within-step depth saturation at the small and medium budgets. Mean return with one standard error across three seeds is reported.}
    \label{fig:appendix-sokoban-convglu-depth-cap}
\end{figure*}

\begin{figure*}[t]
    \centering
    \includegraphics[width=0.98\textwidth]{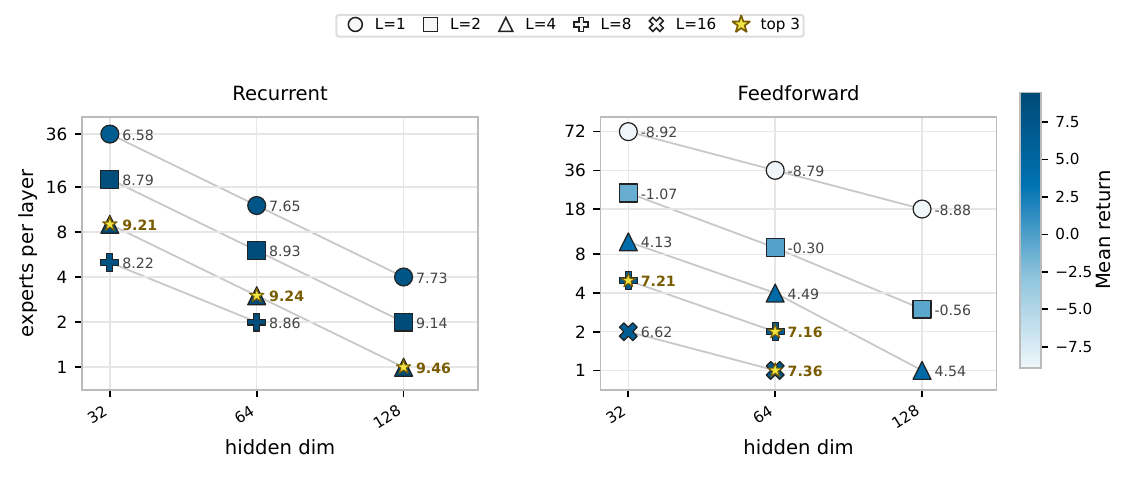}
    \caption{Sokoban small-budget the ConvGLU allocations under recurrent and feedforward settings.}
    \label{fig:sokoban-small-glu-recurrent-feedforward}
\end{figure*}

\begin{figure*}[h!]
    \centering
    \includegraphics[width=0.98\textwidth]{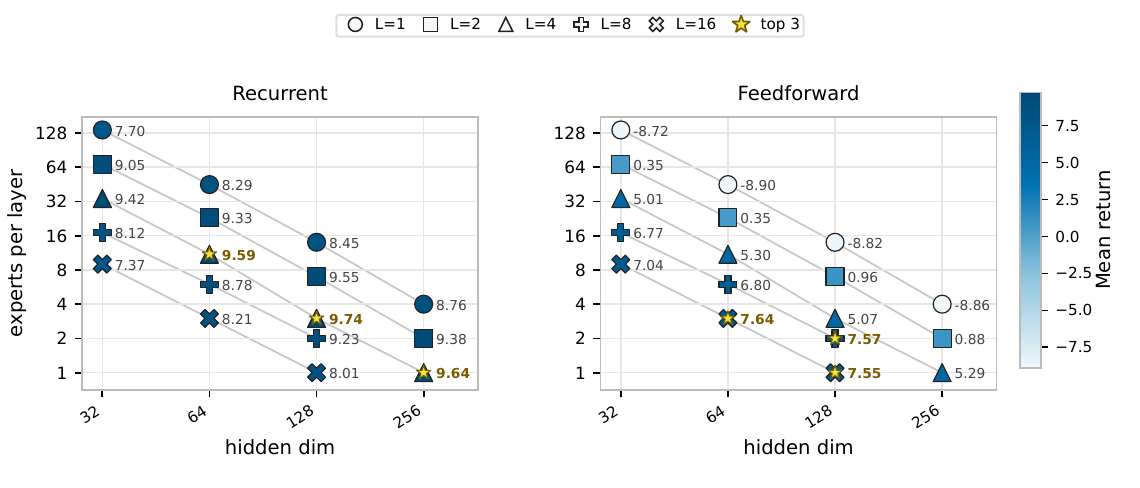}
    \caption{Sokoban medium-budget ConvGLU allocations under recurrent and feedforward settings.}
    \label{fig:appendix-sokoban-medium-convglu-recurrent-feedforward}
\end{figure*}

Fig.~\ref{fig:appendix-sokoban-convglu-depth-cap} shows the similar saturation ans scaling effects but for the ConvGLU architecture in Sokoban. Fig.~\ref{fig:sokoban-small-glu-recurrent-feedforward} and Fig.~\ref{fig:appendix-sokoban-medium-convglu-recurrent-feedforward} show the recurrent and feedforward allocations for the ConvGLU sweeps. 

\FloatBarrier
\clearpage

\end{document}